\documentclass[lettersize,journal]{IEEEtran}
\usepackage{amsmath,amsfonts}
\usepackage{algorithmic}
\usepackage{algorithm}
\usepackage{array}
\usepackage[caption=false,font=normalsize,labelfont=sf,textfont=sf]{subfig}
\usepackage{textcomp}
\usepackage{stfloats}
\usepackage{url}
\usepackage{verbatim}
\usepackage{graphicx}
\usepackage{cite}
\usepackage{bm}
\usepackage{booktabs}
\usepackage{multirow}
\usepackage{tabularx}
\usepackage{cleveref}
\begin{document}

\title{SiamTHN: Siamese Target Highlight Network for Visual Tracking}

\author{Jiahao Bao,
        Kaiqiang Chen,
        Xian Sun,~\IEEEmembership{Senior Member,~IEEE,}
        Liangjin Zhao,
        Wenhui Diao,
        Menglong Yan % <-this % stops a space
\thanks{Corresponding author: Menglong Yan.
% This work was supported by National Key R&D Program of China under Grant No. 2021YFB3900504. 

Jiahao Bao and Xian Sun are with the Aerospace Information
Research Institute, Chinese Academy of Sciences, Beijing
100190, China, 
the Key Laboratory of Network Information System
Technology (NIST), Aerospace Information Research Institute, Chinese
Academy of Sciences, Beijing 100190, China, 
the University of Chinese Academy of Sciences 
and the School of Electronic, Electrical and Communication Engineering, University of Chinese Academy of Sciences,
Beijing 100190, China (e-mail: baojiahao20@mails.ucas.ac.cn; sunxian@aircas.ac.cn).

Kaiqiang Chen, Liangjin Zhao, Wenhui Diao are with the Aerospace Information Research Institute,
Chinese Academy of Sciences, Beijing 100190, China and the Key Laboratory
of Network Information System Technology (NIST), Aerospace Information
Research Institute, Chinese Academy of Sciences, Beijing 100190,
China (e-mail: chenkaiqiang14@mails.ucas.ac.cn; zhaolj004896@aircas.ac.cn; diaowh@aircas.ac.cn).

Menglong Yan is with the Aerospace Information Research Institute, Chinese
Academy of Sciences, Beijing 100190, China, with the Key Laboratory of
Network Information System Technology, Aerospace Information Research
Institute, Chinese Academy of Sciences, Beijing 100190, China, and also with
the Jigang Defence Technology Company, Ltd., Jinan 250132, China (e-mail: yanml@aircas.ac.cn).}
}

% The paper headers
\markboth{Journal of \LaTeX\ Class Files,~Vol.~14, No.~8, August~2021}%
{Shell \MakeLowercase{\textit{et al.}}: A Sample Article Using IEEEtran.cls for IEEE Journals}

% Remember, if you use this you must call \IEEEpubidadjcol in the second
% column for its text to clear the IEEEpubid mark.

\maketitle

\begin{abstract}
Siamese network based trackers develop rapidly in the field of visual object tracking in recent years. The majority of siamese network based trackers now in use treat each channel in the feature maps generated by the backbone network equally, making the similarity response map sensitive to background influence and hence challenging to focus on the target region. Additionally, there are no structural links between the classification and regression branches in these trackers, and the two branches are optimized separately during training. Therefore, there is a misalignment between the classification and regression branches, which results in less accurate tracking results. In this paper, a Target Highlight Module is proposed to help the generated similarity response maps to be more focused on the target region. To reduce the misalignment and produce more precise tracking results, we propose a corrective loss to train the model. The two branches of the model are jointly tuned with the use of corrective loss to produce more reliable prediction results. Experiments on 5 challenging benchmark datasets reveal that the method outperforms current models in terms of performance, and runs at 38 fps, proving its effectiveness and efficiency.
\end{abstract}


\section{Introduction}
\IEEEPARstart{V}{isual} object tracking is a basic challenge with the task of forecasting the target state in each frame of a video. It has several uses in numerous industries, including pose estimation\cite{kart2019object}, person retrieval\cite{shi2020adaptive}, visual surveillance\cite{wu2022pseudo} and autonomous vehicles\cite{gao2019manifold}. Therefore, it is a very active research direction. Despite the recent advances, various issues, such as scale variances, background clutters, scale variation and scale variation, continue to make it a challenging task.

%\begin{figure}[!t]
%\centering
%\includegraphics[width=3.5in]{Baojh1}
%\caption{Tracking comparison between previous work and our work on example frames. The green bounding boxes denote the ground truth, while the tracking results produced by previous work and our work are shown by the blue and red bounding boxes. Clearly, our work achieves significant improvement compared with previous work.}
%\label{Baojh1}
%\end{figure}

\begin{figure*}[!t]
\centering
\includegraphics[width=7in]{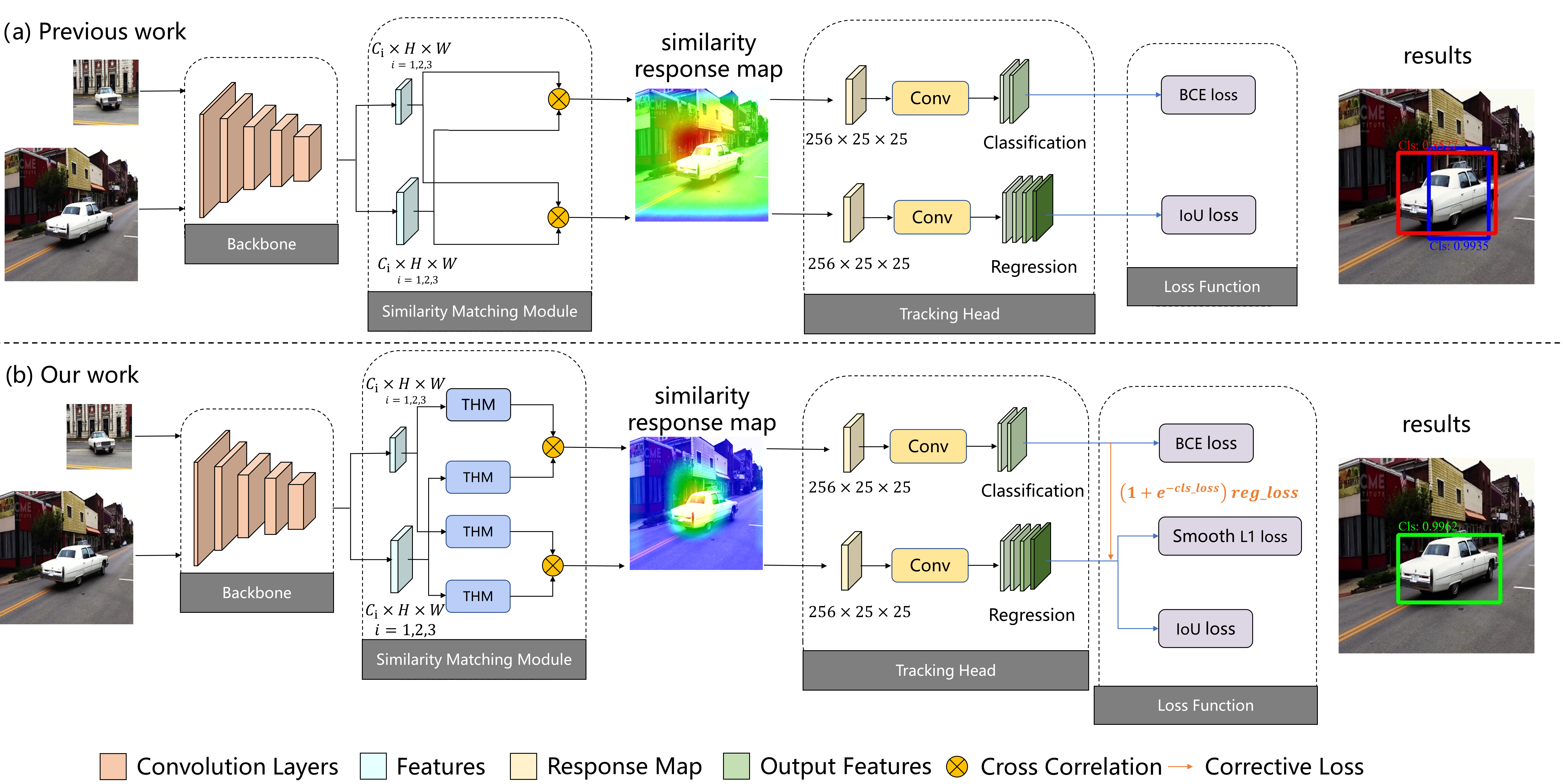}
\caption{Comparison of the previous work and our SiamTHN. We have two innovations. (1) The similarity response map generated by the previous work is easily disturbed by the background. After adding THM, the similarity response map is better focused on the target region. (2) In the previous work, two branches of the tracking head lack a direct connection to each other and are optimized independently. This results in a misalignment between the two branches. As shown in the results of (a), the classification score of the blue bouding box is higher than that of the red bounding box, but the red bounding box generates more accurate tracking results. Training the model with corrective loss can effectively alleviate this problem, so that the generated tracking results can be more accurate.}
\label{Baojh1-1}
\end{figure*}

Correlation filter\cite{henriques2014high}\cite{li2014scale}\cite{kiani2017learning}\cite{han2018adaptive}\cite{zhu2020complementary}\cite{jain2021channel} and siamese network\cite{bertinetto2016fully}\cite{li2019siamrpn++}\cite{chen2020siamese}\cite{fan2020feature}\cite{jiang2020mutual} \cite{fan2021siamon}\cite{wang2021dynamic} are the two popular types of trackers. Wherein, siamese network based trackers\cite{bertinetto2016fully}\cite{li2019siamrpn++}\cite{chen2020siamese} show encouraging results. The pioneering method, SiamFC\cite{bertinetto2016fully} applies the siamese network structure\cite{bromley1993signature} and proposes a cross-correlation layer (Xcorr) for the object tracking issue, establishing the groundwork for a series of later methods. Following this work, although several studies\cite{guo2017learning}\cite{wang2018not}\cite{he2018twofold} focus on ways to enhance the feature representation of the Siamese model, the overall structure has remained mostly unchanged. They are still difficult to solve the scale variation problem of images. Until 2018, SiamRPN\cite{li2018high} introduces region proposal network (RPN)\cite{ren2015faster}, to slove the problem. Since RPN relies on anchor points and a series of related hyperparameters, the model's generalization ability is severely reduced. Therefore, a series of anchor-free trackers are proposed, including SiamBAN\cite{chen2020siamese} and SiamCAR\cite{guo2020siamcar}. In the last two years, transformer\cite{vaswani2017attention} becomes increasingly popular in the field of computer vision, and some work\cite{chen2021transformer}\cite{yan2021learning}\cite{gao2022aiatrack} start to apply it to siamese network based trackers. However, there are two problems in the existing research methods as shown in Fig \ref{Baojh1-1}(a). Firstly, siamese trackers have difficulty in distinguishing background distractors. Specifically, in the process of similarity calculation, the generated similarity response map is difficult to focus on the target region, which will directly affect the effectiveness of feature decoding in the subsequent tracking head. Secondly, the classification branch and the regression branch in the tracking head are separate in processing the task. Specifically, the classification branch is responsible for distinguishing the target from the background, while the regression branch is responsible for locating the bounding box of all positive samples and does not consider the classification information. It results in the accuracy misalignment between the output feature maps. 

On the one hand, most of the methods\cite{li2019siamrpn++}\cite{chen2020siamese}\cite{guo2020siamcar} treat each channel in the feature map equally in the process of channel downscaling, making it difficult to focus the similarity response maps on the target region. DW-Xcorr\cite{li2019siamrpn++} is a common similarity calculation method in the currently popular siamese network based trackers. It convolves the two feature maps extracted by the siamese network channel by channel and outputs the final similarity response map. The similarity response map has a feature that objects of the same category have a higher response on the same channel, while the response of other channels is suppressed. However, existing siamese network based trackers usually use a modified ResNet-50 as the feature extraction network. As a result, the number of channels of the final output feature map is too large, which leads to an elevated computational effort and makes it difficult to meet the real-time requirements of object tracking. In order to reduce the computational effort, they\cite{li2019siamrpn++}\cite{chen2020siamese}\cite{guo2020siamcar} use 1×1 convolution to decrease the feature map's channels. As shown in Fig \ref{Baojh1-1}(a), such processing reduces the computational effort in DW-Xcorr, but leads to difficulty in focusing the similarity response map on the target region.

To address the aforementioned issue, we propose a Target Highlight Module(THM). Existing channel feature balancing methods are generally applied in backbone networks for enhanced feature extraction, but there is still a possibility that critical information will be overlooked in the subsequent dimensionality reduction process. Unlike previous methods, THM performs dynamic channel feature balancing during channel dimensionality reduction to ensure that the feature maps input to DW-Xcorr for similarity calculation have strong target-related features. Therefore, it can strengthen the channels that emphasize the target in the similarity response map while suppressing other unimportant channels in the channel downscaling. The enhanced feature maps obtained by THM are fed into DW-Xcorr, and the resulting similarity response maps can be better focused on the target region. As shown in Fig \ref{Baojh1-1}(b), the similarity response map is more focused on the target region in the presence of THM.

On the other hand, the classification and regression branches in the tracking head are independent of each other, resulting in misalignment between the output feature maps. In siamese network based trackers\cite{li2019siamrpn++}\cite{chen2020siamese}\cite{guo2020siamcar}, there is no direct structural connection between the classification branch and the regression branch and they are optimized independently. However, the regression branch outputs the corresponding prediction bounding box based on the feature map produced by the classification branch during the tracking phase. As a result, there is a large number of inconsistent predictions in the inference stage, which usually have high classification scores but less accurate regression bounding boxes. As shown in results in Fig \ref{Baojh1-1}(a), the blue bounding box has a higher classification score than the red bounding box. Therefore, we output the blue bounding box as the final tracking result. However, the red bounding box is more accurate in terms of tracing results.

To solve this problem, we propose a corrective loss rather than the original loss during the training phase, which can supervise the two branches together towards the optimal direction. Unlike the previous method, corrective loss does not require any additional branches to be added. In corrective loss, the regression loss is specifically corrected by a factor associated with the classification loss. As shown in Fig \ref{Baojh1-1}(b), it multiplies $(1 + e^{-clsloss})$ as coefficients before the regression loss. It is able to converge in a smoother way during model training, thus better facilitating the coordinated optimization of the two branches. Therefore, the two branches produce more consistent predictions, reducing the misalignment problem in the tracking phase. We can see from the results in Fig \ref{Baojh1-1}(b), the point with the highest classification scores also outputs the best bounding box.

Overall, we propose a Target Highlight Module(THM) and a corrective loss to address the shortcomings in existing siamese network based trackers. Specifically, THM is applied in the similarity matching module, which helps the feature map to enhance the features related to the target during the downscaling process, thus making the similarity response map better focus on the target region. 
Corrective loss is used in the training algorithm to establish and strengthen the connection between the classification branch and the regression branch, so as to solve the misalignment between the two branches. Based on these, we develop a model named Siamese Target Highlight Network (SiamTHN). Our proposed THM is a lightweight module and corrective loss enhances the training effect of the model from the perspective of the loss function. Therefore, SiamTHN can achieve good tracking accuracy while maintaining high fps, which is more suitable for real-world scenarios. As shown in Fig \ref{Baojh1-1}, the tracking frame produced by our model is significantly more accurate compared to the previous model. To sum up, the contributions of this paper can be summarized in the following three aspects.

\begin{enumerate}
\item{We propose a Target Highlight Module(THM) which helps the similarity response map to be more focused on the target.}
\item{We propose a corrective loss to optimize the regression loss using classification loss supervision during training phase, which can alleviate the misalignment between two branches. As a result, the model is able to produce predictions that are more accurate.}
\item{On the basis of THM, we propose a Siamese Target Highlight Network (SiamTHN). Experiments on multiple challenging benchmark datasets show that Siamese Target Highlight Network (SiamTHN) perform better than several state-of-the-art trackers and achieves leading performance.}
\end{enumerate}

The remainder of this article is organized as follows. Section II reviews related work in three parts: siamese networks based tracker, channel attention and bounding box localization strategy. Section III describes the overall framework of the Siamese Target Highlight Network (SiamTHN), Target Highlight Module, and the corrective loss. Section IV presents a qualitative and quantitative experimental evaluation of our method compared to other state-of-the-art trackers. Additionally, we conduct ablation experiments as a way to prove the effectiveness of the our module and loss, and to quantitatively evaluate their separate contributions. Finally, in Section V, conclusions are formed.

%\begin{figure*}[!t]
%\centering
%\includegraphics[width=7in]{Baojh2}
%\caption{Network architecture of our proposed Siamese Target Highlight Network (SiamTHN). It starts with a siamese network as backbone, which is used to extract features. Then, there is a similarity matching module with our novel Target Highlight Module(THM). It is used to produce similarity response maps that assess how similar the search and template images are to one another. Finally, there are classification branch and regression branch, which together are called tracking head. Tracking head is responsible for feature decoding of the output similarity response map to produce the final outputs. In addition, the far right side of the figure shows how our corrective loss works.}
%\label{Baojh2}
%\end{figure*}

\section{Related Work}
In this section, we will concentrate on the following three aspects that are most relevant to our work, including siamese network based trackers, attentional mechanisms and bounding box localization strategy.

\subsection{Siamese Network based Trackers}
In recent years, siamese network based trackers\cite{bertinetto2016fully}\cite{li2019siamrpn++}\cite{chen2020siamese}\cite{fan2020feature} \cite{jiang2020mutual} \cite{fan2021siamon}\cite{wang2021dynamic} achieves a great deal of breakthroughs in visual object tracking. These trackers share a lot of structural similarities, which consist of a siamese network, a similarity matching module, and a tracking head. Naturally, most of the research has concentrated on optimizing and improving these three components, as shown in Fig \ref{Baojh1-1}.

As one of the pioneering works, SiamFC\cite{bertinetto2016fully} introduces siamese network to visual target tracking for the first time\cite{li2018deep}. It modifies and builds the network on top of AlexNet\cite{krizhevsky2012imagenet} to extract features. For better application, the siamese network removes the padding and fully connected layers and adds a batch normalization layer. In addition, SiamFC proposes a cross-correlation layer (Xcorr) for the correlation operation of template features and search features. Specifically, the template feature map is used as a convolution kernel to convolve with the search feature map, which produces the similarity response map. In essence, it contains a information of the similarity between the template and the search region. Then, the researchers go on to build some revised siamese methods\cite{guo2017learning}\cite{wang2018not}\cite{he2018twofold} on the basis of this siamese framework. DSiam\cite{guo2017learning} proposes dynamic siamese networks which can learn target appearance changes and background suppression. RASNet\cite{wang2018not} introduces spatial attention and channel attention mechanisms. However, these trackers are all based on SiamFC's framework, which means they can only achieve multi-scale search by inputting images of multiple scales to deal with scale variation.

Then, SiamRPN\cite{li2018high} introduces the region proposal network (RPN)\cite{ren2015faster} to the siamese network based trackers. The RPN is made up of two branches: a classification branch and a regression branch. The regression branch is used to regress the bounding box, while the classification branch is used to distinguish between the target's foreground and background. In addition, SiamRPN also introduces the up-channel cross correlation layer (Up-Xcorr). It outputs a multi-channel similarity response map which is sent to RPN for feature decoding. As a result, the typical multi-scale search can be discarded, greatly increasing the speed of inference. After that, SiamRPN++\cite{li2019siamrpn++} deepens the siamese network. It removes the stride from the last two blocks of ResNet\cite{he2016deep} and adds the dilated convolution\cite{long2015fully}. And the modified ResNet is applied to the feature extraction network in siamese network based trackers. Apart from this, SiamRPN++ also proposes a depth-wise cross correlation layer (DW-Xcorr). DW-Xcorr, in comparison to Up-Xcorr, solves the problem of imbalanced parameter distribution in the two branches while drastically reducing the number of parameters. SiamRPN++ is more consistent during the training process, and its performance is significantly improved. 

SiamRPN++ refines the basic framework of siamese network based trackers, and most of the subsequent trackers are improved with this framework. GradNet\cite{li2019gradnet} proposes that the existing framework template is fixed with the initial target features and the performance is completely dependent on the overall matching ability of the siamese network. Therefore, it proposes a template generalization training method using gradient information for template updating. There are also algorithms that focus on the shortcomings in RPN and improve them. C-RPN\cite{fan2019siamese} proposes to solve the class imbalance problem by cascading a series of RPNs in a siamese network from deep layers to shallow layers. Some other studies concludes that RPN must rely on a huge number of hyperparameters related to the anchors, which considerably decreases the tracker's generalization performance. Furthermore, the scale and aspect ratio of anchor box are fixed and require strong priori knowledge to design. In order to solve these problems, the anchor-free method is proposed, such as SiamFC++\cite{xu2020siamfc++}, SiamBAN\cite{chen2020siamese}, SiamCAR\cite{guo2020siamcar} and Ocean\cite{zhang2020ocean}. They use a per-pixel-prediction method to regress the bounding box from the similarity response map. In this way, they can get rid of the inconvenient anchor hyperparameters. As the performance of siamese network based trackers continues to improve, some algorithms\cite{ramesh2020tld} start to focus more on long-term object tracking and some other more practical application scenarios. 

However, the similarity response maps generated by existing siamese network based trackers\cite{li2019siamrpn++}\cite{chen2020siamese} do not focus well on the target, and the classification branches and regression branches are optimized independently. Different from the previous tracker, we propose a novel channel attention module called Target Highlight Module which can highlight the similarity response map's target region.  Additionally, we suggest that the model trained with our corrective loss can more effectively reduce the misalignment between the feature maps produced by two branches. The related work about Target Highlight Module and corrective loss is reviewed in section B and section C.

\begin{figure}[!t]
\centering
\includegraphics[width=3.5in]{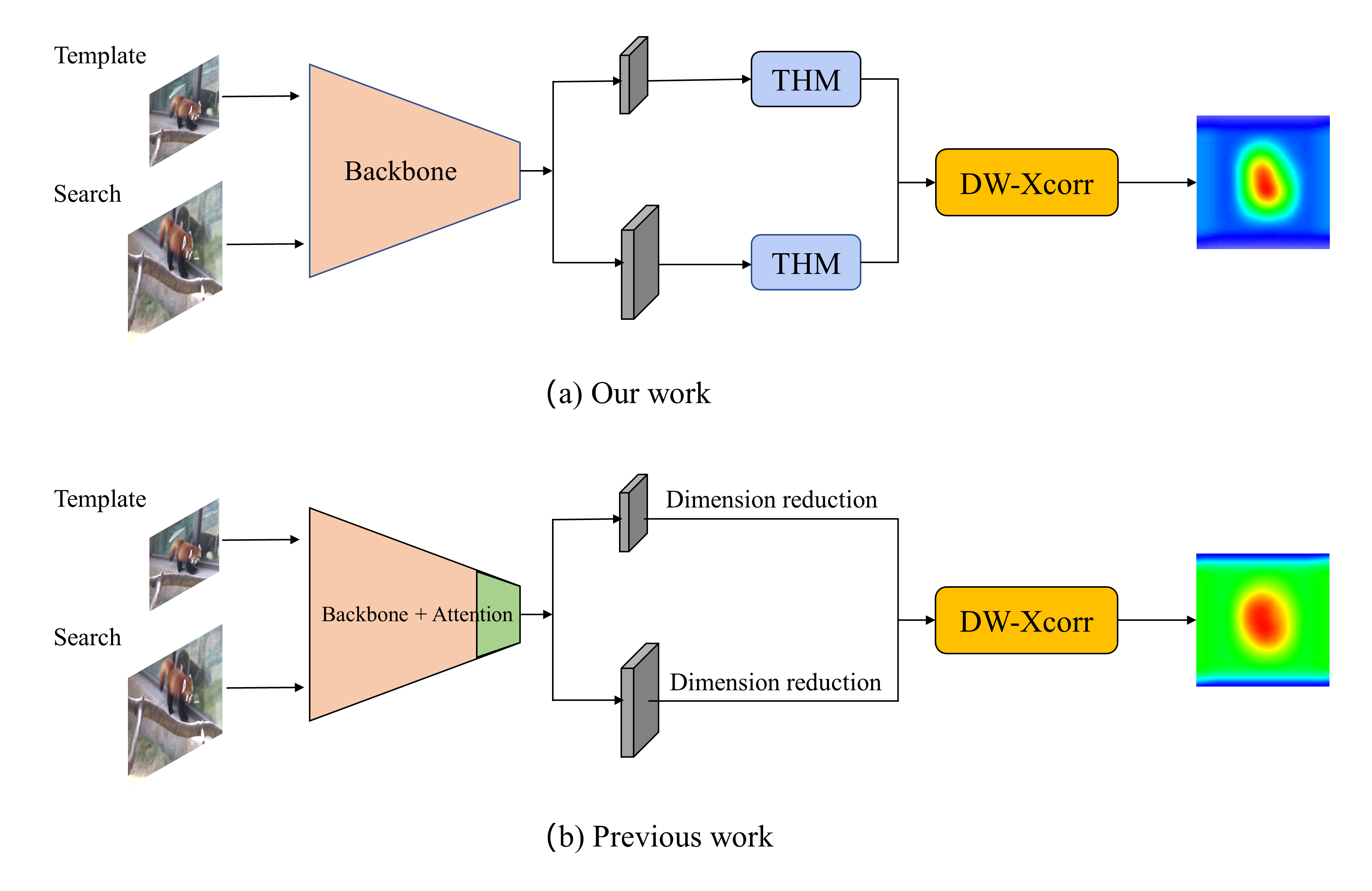}
\caption{Comparison of the our work and previous work. In the previous work, attention mechanism is used to improve the feature extraction ability of backbone. In our work, Target Highlight Module is located in the similarity matching module, which is used to help the feature map to perform better feature selection and enhancement during the dimensionality reduction process.}
\label{Baojh3-1}
\end{figure}

\subsection{Attentional Mechanisms}
Attention mechanism can be described as an algorithm for dynamic weight modification based on the input image features. It excels at a variety of visual tasks since its debut, including image classification and object detection. In deep neural networks, different channels in different feature maps usually represent different objects\cite{chen2017sca}. As the first approach, SENet\cite{hu2018squeeze} introduces the concept of channel attention and the squeeze-and-excitation (SE) block. The core idea is to collect global spatial information using global average pooling, then output the channel attention map using non-linear activation functions and fully-connected layers. However, SENet also has many drawbacks. In the squeeze module, global average pooling is difficult to capture complex global information. In the excitation module, fully-connected layers increase the complexity of the model. Subsequent improvements are also focused on these two modules. GSoP-Net\cite{gao2019global} is dedicated to enhance the modeling capability of squeeze module. On top of the basic global average pooling, it introduces a global second-order pooling to model higher-order statistics. ECANet\cite{2020ECA} works to reduce the complexity of the excitation module by useing a 1D convolution to determine the interaction between channels. Later, SRM\cite{lee2019srm} is inspired by style transfer to improve both the squeeze module and the excitation module. It introduces style pooling, which enhances the acquisition of global information by using both the mean and standard deviation of the input features. Additionally, instead of the original fully-connected layer, it suggests a channel-wise fully-connected layer to reduce the computational requirements. Recently, transformer\cite{vaswani2017attention} becomes increasingly popular in the field of computer vision. Its model structure is based entirely on the attention mechanism without any convolutional or recurrent neural network layers.

Attention is also used in the field of visual object tracking. RASNet\cite{wang2018not} introduces the attention mechanism proposed by SENet to siamese network based trackers for the first time. It mainly uses the attention mechanism to enhance the representation of feature maps. SA-Siam\cite{he2018twofold} suggests calculating channel direction weights based on channel activation at the target location. The previous work mainly use channel attention to enhance the output feature maps. FAliM\cite{fan2020feature} aggregates shallow and high level features and uses the channel attention mechanism to enhance the discriminative power of the aggregated feature representation. TransT\cite{chen2021transformer} suggests an unique attention-based feature fusion network based on transfomer inspiration. And AiATrack\cite{gao2022aiatrack} proposes an attention-in-attention (AiA) module.

\begin{figure}[!t]
\centering
\includegraphics[width=3.5in]{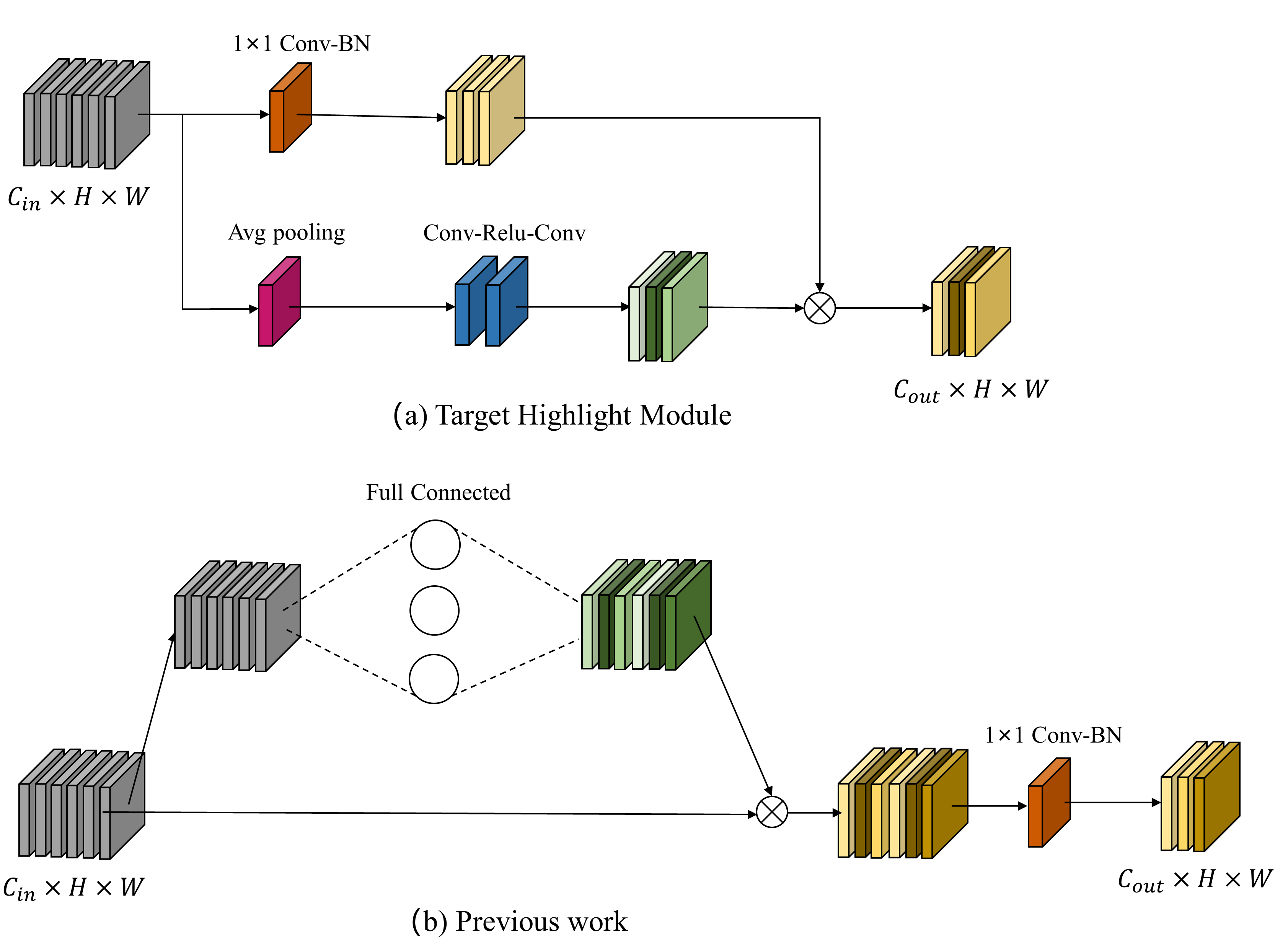}
\caption{Illustration of the Target Highlight Module(THM). (a) shows the operation of Target Highlight Module. (b) shows the operation of previous work. The previous work uses a fully connected layer to enhance the feature map of the backbone network output, but still lose important channel information in the subsequent channel downscaling process. THM is used during the channel downscaling process, and the use of convolutional layers can better preserve the position and shape information in the feature map.}
\label{Baojh3-2}
\end{figure}

Our proposed Target Highlight Module(THM) is a kind of channel attention, but differs from the previous work: (1) The motivations are different. THM is proposed to help input feature maps of DW-Xcorr for better feature selection during channel downscaling. Previous channel feature balancing methods mainly focus on solving the feature enhancement problem for backbone networks. Therefore, they are applied to different locations in siamese network based trackers. Fig \ref{Baojh3-1} is a simple schematic diagram showing the process of generating similarity response map in siamese network based trackers. The process is as follows: two input images are extracted by backbone network and then input to similarity matching module. As shown in Fig \ref{Baojh3-1}, THM is located in the similarity matching module, while previous channel feature balancing methods are located in backbone network. (2) The implementations are different. As shown in Fig \ref{Baojh3-2}, THM uses convolutional layers to learn the spatial structure properties and channel weights of the feature maps in the process of channel dimensionality reduction. However, previous channel feature balancing methods, although using full connected layers to enhance the feature map of the backbone network output, may still lose important channel information in the subsequent channel downscaling process. (3) The problems solved are different. THM and the previous channel feature balancing methods solve different problems, and directly transposing the previous channel feature balancing methods into the similarity matching module does not work well. We conduct an experiment to demonstrate this in the experiments section.

\subsection{Bounding Box Localization Strategy}
In most of the object detectors\cite{liu2016ssd}\cite{girshick2015fast}\cite{ren2015faster}, two parallel head structures are widely used to handle the classification and regression tasks separately. However, model's prediction result can be considerably affected by spatial misalignment between the output feature maps of the two branches. IoU-Net\cite{jiang2018acquisition} is the first to reveal this problem and and proposes to use the predicted IoU as localization confidence. PISA\cite{cao2020prime} proposes a Classification-Aware Regression Loss (CARL), in which samples with higher regression losses have higher classification score gradients. As a result, regression loss is able to supervise the optimization of classification branch. In recent, Harmonic loss\cite{wang2021reconcile} proposes classification and regression branches can supervise each other's optimization during training to produce consistent prediction results in the inference phase.

The similar difficulty emerges in target tracking since the design of the head of many trackers is based on the head of detectors. SiamCAR\cite{guo2020siamcar} and SiamFC++\cite{xu2020siamfc++} estimates the bounding box quality by introducing an additional branch\cite{tian2019fcos}. SiamRCR\cite{peng2021siamrcr} proposes a method to fuse classification loss and regression loss. However, it still needs to add an additional localization branch to predict the localization accuracy. We want to be able to solve this problem purely using loss functions without adding additional branches. Therefore, we propose corrective loss. Although our corrective loss shares partial similarity with the above methods, the technical details are quite different. (1) The corrective loss does not require additional branches to be added to the model. (2) We fuse Smooth L1 loss\cite{girshick2015fast} and IoU loss\cite{yu2016unitbox} to make our regression loss more stable and reliable. (3) We use $(1 + e^{-clsloss})$ as the coefficient of regression loss to match the final inference process. The advantage of doing this is that the loss function converges better. For a positive sample, if its classification score is lower, the corresponding regression loss weight will be smaller. To better illustrate the validity of corrective loss, we set up a set of experiments.

\section{Method}
This section focuses on the new tracking model, which is developed based on SiamBAN\cite{chen2020siamese}. We propose a Target Highlight Module and corrective loss function to improve it, and our model is called Siamese Target Highlight Network (SiamTHN).  First, we make a introduction of our proposed SiamTHN. Then, we will illustrate the architecture of Target Highlight Module. Finally, corrective loss function are described in detail.

\subsection{Siamese Target Highlight Network}
Our baseline model, SiamBAN is a simple yet effective Siamese tracking framework. However, there are some issues with the model's design. First, SiamBAN treats each channel equally when performing channel downscaling on the feature map, ignoring the fact that various channels respond differently to the target. Second, the classification and regression branches are optimized independently in the SiamBAN, resulting in misalignment between them. Based on these two aspects, we develop the Siamese Target Highlight Network (SiamTHN) to improve SiamBAN.

Our Siamese Target Highlight Network (SiamTHN) is mainly built on the SiamBAN architecture. As shown in Fig \ref{Baojh1-1}(b), Siamese Target Highlight Network consists of siamese network, similarity matching module, and tracking head. Siamese network is mainly responsible for extracting features from the template and search images. We build the siamese network around a modified ResNet-50\cite{he2016deep} and extracted the feature maps from the last three blocks. To produce multi-channel similarity response maps, the resulting feature maps are passed into the similarity matching module. As shown in Fig \ref{Baojh1-1}(b), our proposed Target Highlight Module(THM) is added to the similarity matching module. Specifically, it helps input feature maps for better feature selection during channel downscaling. Therefore, similarity matching module can produce higher quality similarity response maps. Finally, the similarity response maps would be fed into the tracking head, which consists of a classification branch and a regression branch. The classification branch is responsible for decoding features of the input similarity response map to generate classification score map, while the regression branch is responsible for outputting the bounding box regression map. However, there is no structural connection between the classification branch and the regression branch, and their respective tasks are performed independently. Therefore, we suggest using corrective loss to resolve the misalignment between the two branches to produce more consistent prediction results.

\subsection{Target Highlight Module}
In the visual object tracking task, measuring the similarity between the template and the search region is a crucial step. Most advanced methods\cite{li2019siamrpn++}\cite{chen2020siamese}\cite{guo2020siamcar} tend to use the network structure of ResNet-50 in siamese network and the output feature maps are passed through DW-Xcorr to obtain the similarity maps. However, the last three blocks of ResNet-50 produce feature maps of 512, 1024, and 2048 channels. If these feature maps are not processed and sent directly into DW-Xcorr, the computational effort required is enormous and unmanageable. As a result, most of the methods\cite{li2019siamrpn++}\cite{chen2020siamese}\cite{guo2020siamcar} ccrop the 7×7 center region of the feature map and decrease the output feature channels to 256 by using 1×1 convolution. It can be thought of a process of information compression, 
and different channels actually represent different semantics in the feature maps produced by the DW-Xcorr. That is, the importance of the information provided in different channels is not same to the target. The existing methods\cite{li2019siamrpn++}\cite{chen2020siamese}\cite{guo2020siamcar} ignore the fact and treat each channel equally throughout the channel downscaling process, resulting in a feature map generated by DW-Xcorr that is difficult to focus on the target region. Specifically, the previous method generates a similarity response map that is not well focused on the target region but is sensitive to the background region. 

We propose a Target Hightlight Module (THM) to perform better channel selection when downscaling the feature map. As shown in Fig \ref{Baojh3-2}(a), we start by using average pooling to compress the input feature map's spatial dimension. Average pooling gives more accurate feedback for each pixel point on the feature map when it is compressed in spatial dimensions. After that, the feature map is compressed to 1/8 of the original number of channels by a convolution layer to obtain global features at the channel level. The responsiveness of different channels to the target is then learned, and a convolutional layer is utilized to extend it to the output channel in order to obtain the weights of different channels. In contrast to the previous approach, we use a convolutional layer here, which is able to correctly retain and understand the position and shape information of the feature map compared to the full connection layer. For different tracking targets, different channels have variable responsiveness, and Target Highlight Module gives the channels with higher responsiveness and larger weights. Therefore, THM can help the feature map to better enhance the feature representation for the target when downscaling. As a result, the feature map generated by the Target Highlight Module is sent into DW-Xcorr, which generates a similarity response map that is more focused on the target's location.

The specific formula is as follows, given the input feature tensor $f \in R^{H \times W \times C_{in}}$, the equation for dimensionality reduction using channel attention is:
\begin{equation}
\label{deqn_ex1}
 W(f) = \pi(f) \cdot conv(f),
\end{equation}
where $\pi$ is the weight of the feature map after dimensionality reduction. $conv$ denotes the convolutional layer for dimensionality reduction.

Then, the weight of the feature map can be expressed as:
\begin{equation}
\label{deqn_ex2}
 \pi(f) = \delta(conv2(conv1(avg(f)))),
\end{equation}
where $avg$ denotes average pooling, $conv1$ and $conv2$ denote the convolutional layers, and $\delta$ denotes sigmoid function. 

\subsection{Corrective loss}
A tracking head typically has two branches: a classification branch that produces the classification score map and a regression branch that produces the bounding box of the regression. Besides, they choose the bounding box with maximum classification confidence as the final prediction during the tracking phase. However, among the existing siamese trackers, the classification and regression branches are optimized separately and do not have a direct structural link. This results in a misalignment between the two branches. SiamCAR\cite{guo2020siamcar} and SiamFC++\cite{xu2020siamfc++} propose centerness to alleviate this problem. SiamRCR\cite{peng2021siamrcr} propses a reciprocal relationship to solve it.However, each of these methods requires additional branches to be added to the original model. And the challenge of independent optimization of the classsification and regression branches is not solved well. For this purpose, corrective loss is devised.

\begin{figure*}[!t]
\centering
\includegraphics[width=7in]{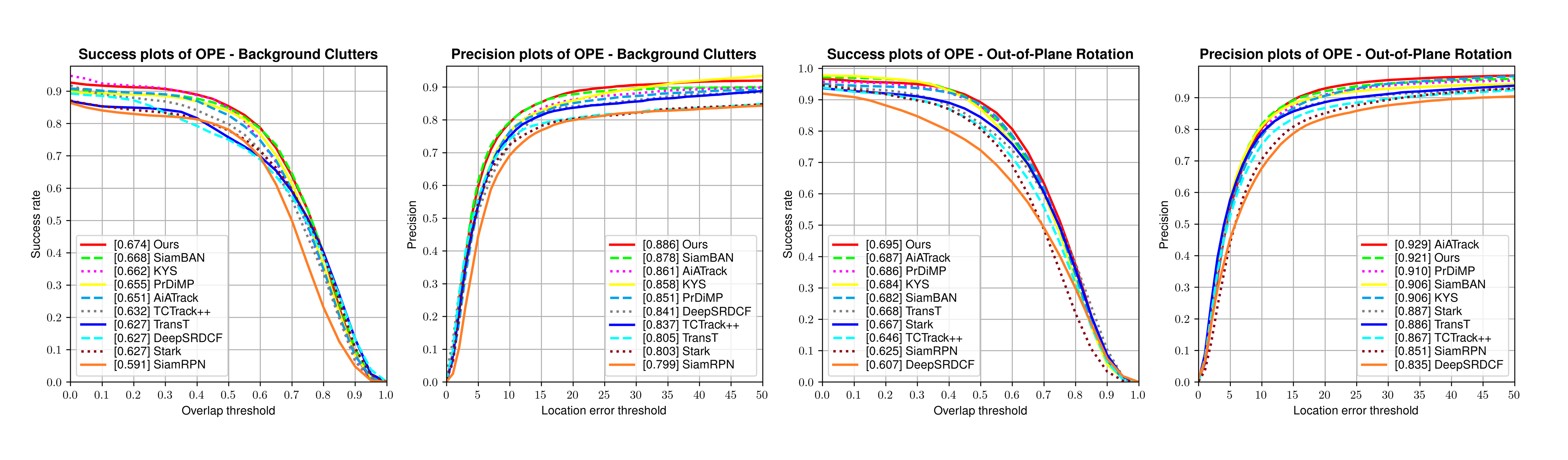}
\caption{Comparisons on OTB-2015 with challenging aspects. Our tracker achieves state-of-the-art performance on two challenges, Background Clutters and Out-of-Plane Rotation.}
\label{Baojh4}
\end{figure*}

\begin{figure}[t]
\raggedright 
\includegraphics[width=3.5in]{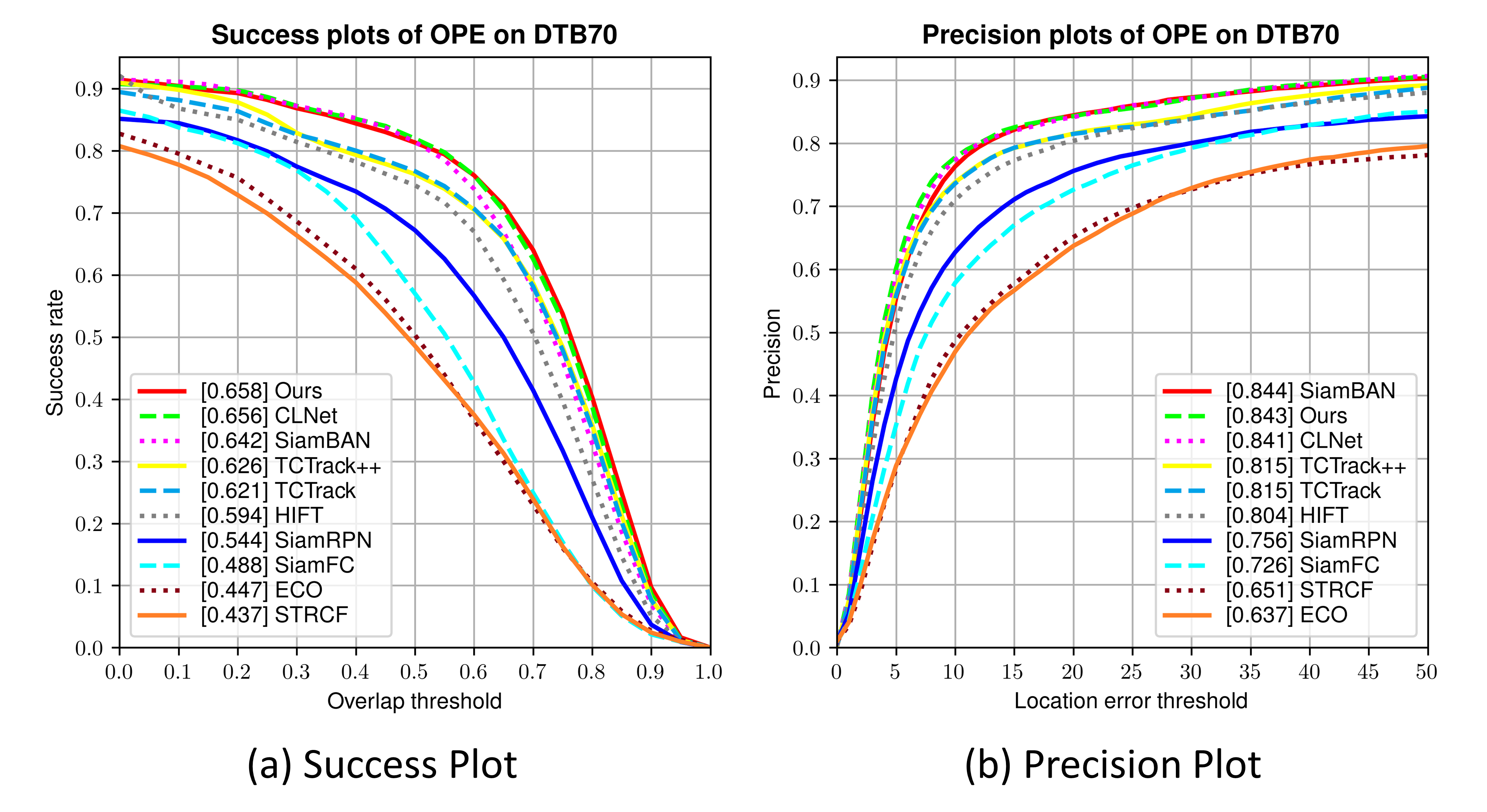}
\caption{Tracking results on DTB70. Our tracker achieves state-of-the-art performance on DTB70 dataset.}
\label{Baojh5}
\end{figure}

In object tracking, the IoU-based localization loss\cite{yu2016unitbox} is denoted as:
\begin{equation}
\label{deqn_ex3}
 L_{IoU} = 1 - IoU,
\end{equation}
where $IoU$ denotes Intersection over Union between prediction results and ground truth. 

However, IoU loss is not perfect. The value of IoU would be zero if is no overlap between the bounding box and the ground truth. The gradient of the loss function is now zero, making it impossible to optimize the parameters. IoU loss can not satisfy our requirement for regression loss in this situation. As a result, we add Smooth L1 loss\cite{girshick2015fast} in addition to the initial IoU loss. Smooth L1 loss can evaluate the regression offset of the output by computing the distance between each feature map point and four edges in the ground truth. As a result, we merge the Smooth L1 loss and IoU loss to produce the regression branch's loss function. The specific formula is as follows:
\begin{equation}
\label{deqn_ex4}
 L_{reg} = L(d, \hat{d}) + L_{IoU},
\end{equation}
where $d$ stands for the output regression offset and $\hat{d}$ represents the target offset. $L$ denotes Smooth L1 loss.

As we mention above, the optimization between the two branches in the past approach was independent. By watching the tracking phase of existing methods, we find that most of them using classification branches to drive regression branches. As a result, we devise a weighted loss function that assigns a different weight to regression loss based on the classification score. The classification loss utilizes cross entropy loss, while it is used to correct the regression loss. Our modifications are mainly focused on the loss function for positive samples. Therefore, the loss function for positive samples $x_i$ is as follows:
\begin{equation}
\label{deqn_ex5}
 L_{pos} = CE(p_i, y_i) + (1 + e^{-CE(p_i, y_i)})L_{reg},
\end{equation}
where $p_i$ stands for the predicted classification score and $y_i$ represents the ground truth class. $CE$ denotes cross entropy loss.

The above equation demonstrates how the classification loss affects the regression loss. Specifically, we multiply a coefficient related to the classification loss in front of the regression loss, allowing the two branches to establish a connection. The regression loss can perceive the classification loss during the training phase. Namely, a positive sample with higher classification score will receive large weight of the regression loss. As a consequence, the two branches will produce more consistent prediction outputs throughout the inference phase, resulting in extraordinarily high localization accuracy.

After adding negative samples, the overall loss is as follows:
\begin{equation}
\label{deqn_ex6}
 L_{all} = \frac{1}{N}(\sum_{i \in pos}^N L_{pos} + \sum_{j \in neg}^M CE(p_j, y_j)).
\end{equation}

\section{Experiments}
In this section, we perform a comprehensive experimental evaluation of the our Siamese Target Highlight Network (SiamTHN). Experiments are conducted on five tracking benchmarks, OTB-2015\cite{7001050}, VOT2016\cite{kristan2016novel}, DTB70\cite{li2017visual}, UAV123\cite{mueller2016benchmark} and UAV20L\cite{mueller2016benchmark}. First, we present the datasets used in the experiments and the implementation details of the training process. Next, five tracking benchmarks and the corresponding evaluation metrics are presented. Finally, we present the ablation experiments and studies about the new module.

\begin{figure}[t]
\raggedright 
\includegraphics[width=3.5in]{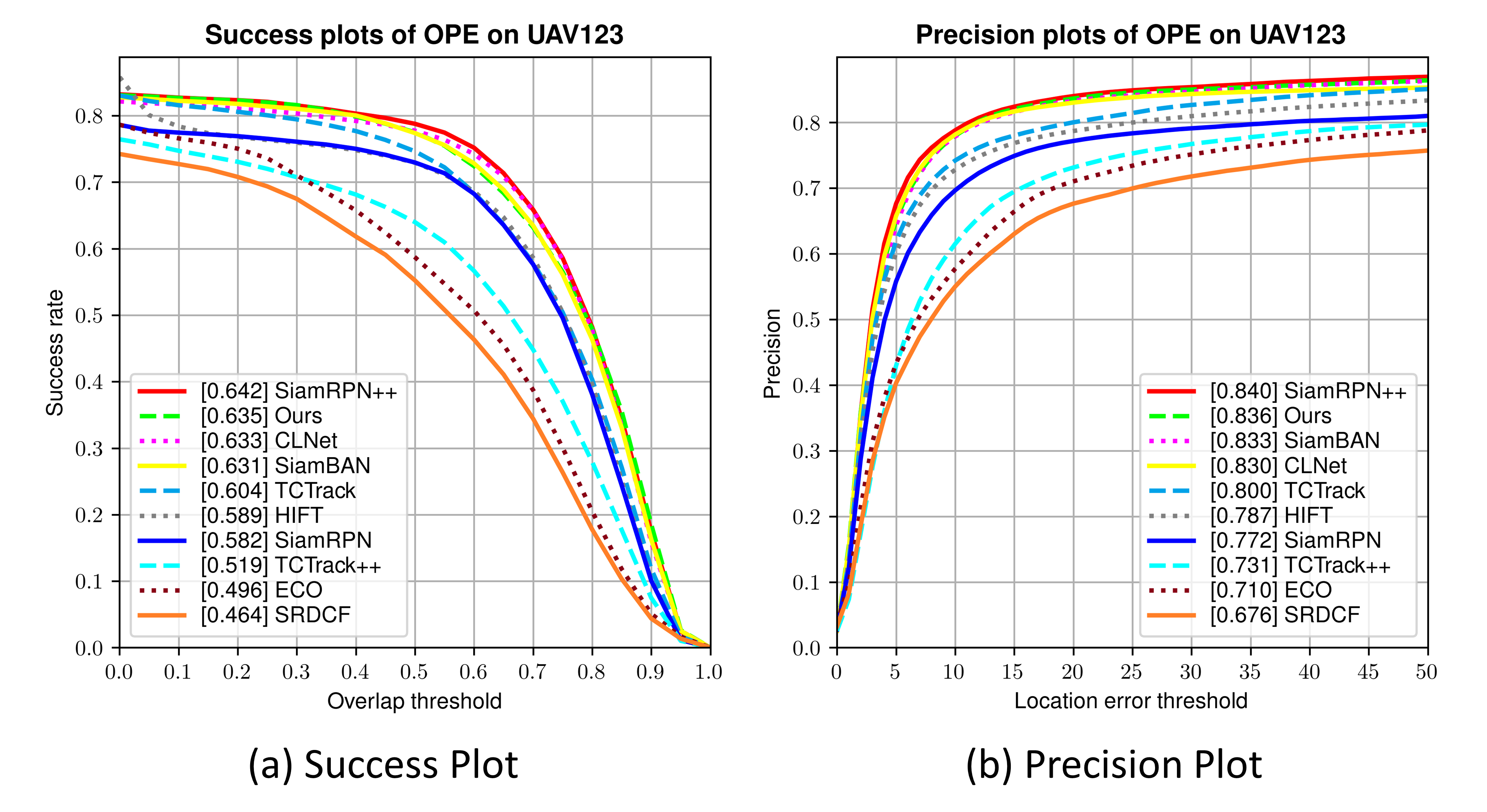}
\caption{Tracking results on UAV123. Our tracker achieves state-of-the-art performance on UAV123 dataset.}
\label{Baojh6}
\end{figure}

\subsection{Dataset}
In this research, we use GOT-10k\cite{huang2019got}, COCO\cite{lin2014microsoft}, ImageNet VID\cite{russakovsky2015imagenet} and ImageNet DET\cite{russakovsky2015imagenet} to train our Siamese Target Highlight Network (SiamTHN). On several well-known tracking benchmarks, including OTB-2015\cite{7001050}, VOT2016\cite{kristan2016novel}, DTB70\cite{li2017visual}, UAV123\cite{mueller2016benchmark} and UAV20L\cite{mueller2016benchmark}, we test our model. OTB-2015 and VOT2016 are two classic single object tracking datasets, and testing on these two datasets can better judge the performance of our method. DTB70, UAV123 and UAV20L are three UAV aerial photography datasets. They contain more challenges and are closer to real application scenarios. Therefore, they can better illustrate our method's application value. We will first give a quick overview of these datasets.

\textbf{GOT-10k}\cite{huang2019got} contains 10,000 videos with over 1.5 million manually annotated bounding boxes. It is built based on the backbone of WordNet structure\cite{miller1995wordnet}, which is used to ensure the category balance in the videos.

\textbf{COCO}\cite{lin2014microsoft} is a large-scale dataset that can be used for a variety of image tasks. It has more than 330K images, 220K of which are annotated, and contains 1.5 million targets, 80 target classes, and 91 material classes.

\textbf{ImageNet}\cite{russakovsky2015imagenet} consists of 14,197,122 images and is a large computer vision dataset. It has many sub-datasets with different divisions. Among them, ImageNet VID has a total of 30 categories, which is a subset of 200 categories of ImageNet DET dataset.

\textbf{OTB-2015}\cite{7001050} consists of 100 videos of 22 object categories. It also
defines 9 attributes such as \textit{Scale Variation}, \textit{Out-of-Plane Rotation}, \textit{Occlusion} and \textit{Deformation}. The video length of OTB-2015 dataset varies from 71 to 3872 frames, with an average resolution of 356×530.

\textbf{VOT2016}\cite{kristan2016novel} contains 60 sequences. Each sequence is labeled by different attributes for each frame, including IV, MOC, SCO, ARC, OCC, and FCM. Sequences typically have a resolution of 757 x 480, with frame sizes ranging from 48 to 1507 pixels.

\textbf{DTB70}\cite{li2017visual} is a dataset which contains 70 video sequences with RGB data. The robustness of the tracker in fast motion scenes may be properly assessed on this benchmark since these sequences feature a significant number of severe motion scenarios. The original resolution of each sequence is 1280 × 720.

\textbf{UAV123}\cite{mueller2016benchmark} is an aerial video benchmark with 123 sequences captured from low-altitude aerial views in its dataset. The benchmark can be used to determine whether the tracker is appropriate for deployment on a UAV in practical situations. It has 123 brief sequences of 9 different object categories, with a minimum frame count of 109 and a maximum frame count of 3085.

\textbf{UAV20L}\cite{mueller2016benchmark} is made up of 20 long videos showing 5 different object classes that were produced using a flight simulator. These sequences have a minimum frame count of 1717 and a maximum frame count of 5527.

\subsection{Evaluation criteria}
For OTB-2015\cite{7001050}, DTB70\cite{li2017visual}, UAV123\cite{mueller2016benchmark} and UAV20L\cite{mueller2016benchmark}, we use the precision plot and success plot to evaluate the performance of the tracker.

\begin{figure}[t]
\raggedright 
\includegraphics[width=3.5in]{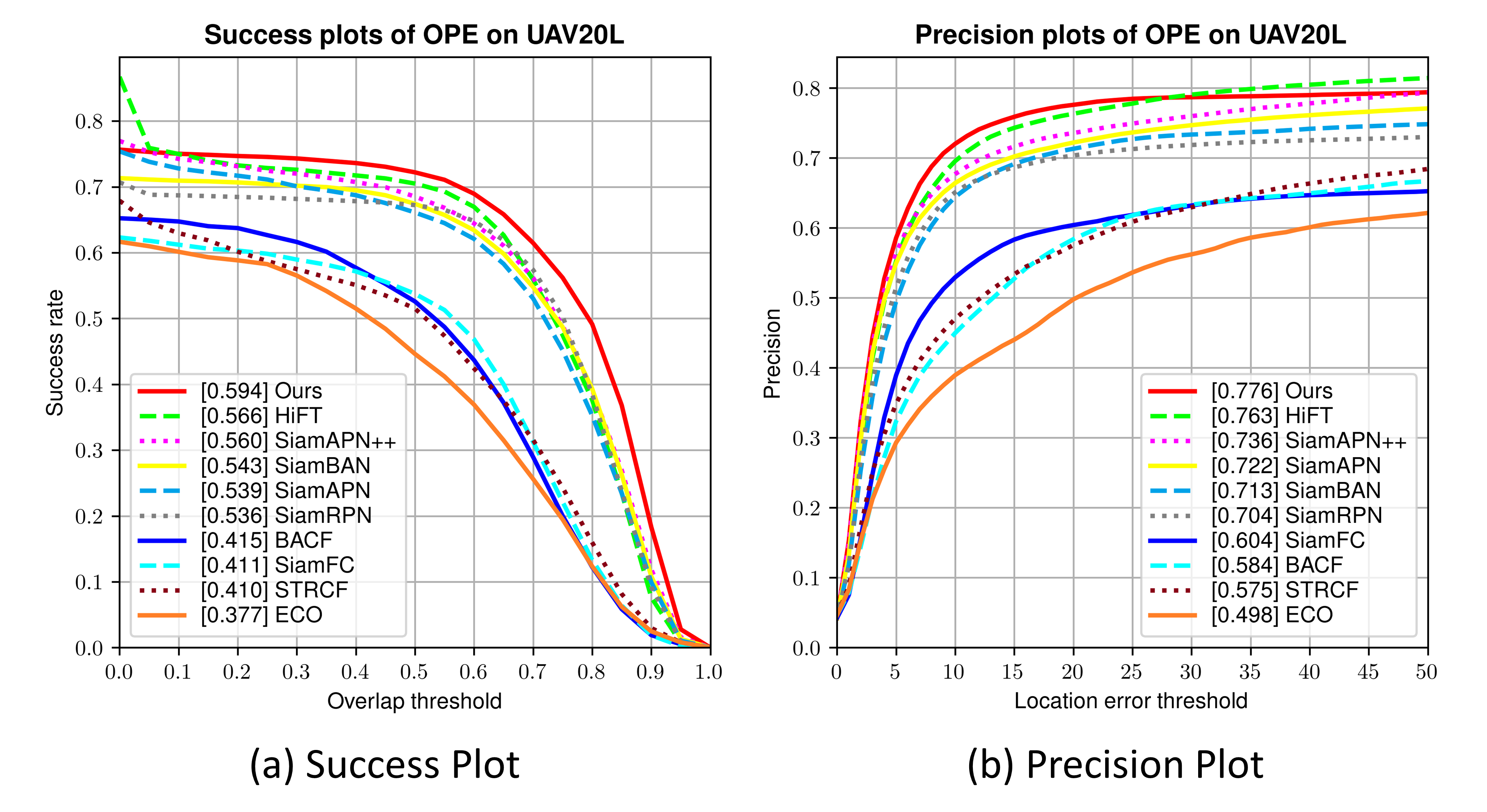}
\caption{Tracking results on UAV20L. Our tracker achieves state-of-the-art performance on UAV20L dataset.}
\label{Baojh7}
\end{figure}

\begin{table}[t]
\caption{TRACKING RESULTS ON VOT2016 DATASET\label{tab:table1}}
\centering
\begin{tabular}{cccc}
\toprule
   \textbf{Tracker} & \textbf{EAO} & \textbf{Accuracy} & \textbf{Robustness}\\
   \midrule
SiamRPN\cite{li2018high} & 0.344 & 0.560 &1.12\\
SiamRPN++\cite{li2019siamrpn++} & 0.370 & 0.580 & 0.240\\
ECO\cite{danelljan2017eco} & 0.374 & 0.546 & 11.67\\
ATOM\cite{danelljan2019atom}& 0.424 & 0.617 & 0.190\\
SiamR-CNN\cite{voigtlaender2020siam} & 0.461 & 0.645 & 0.173\\
PrDiMP\cite{danelljan2020probabilistic} & 0.476 & \textbf{0.652} & 0.140\\
SiamBAN\cite{chen2020siamese} & 0.505 & 0.632 & 0.150\\
Ours & \textbf{0.510} & 0.625 & \textbf{0.126}\\
\bottomrule
\end{tabular}	
\end{table}

\begin{table}[t]
\caption{ANALYSIS OF COMPUTATIONAL COMPLEXITY\label{tab:table2}}
\centering
\begin{tabular}{cccc}
\toprule
   \textbf{Trackers} & \textbf{Flops(G)} & \textbf{Params(M)} & \textbf{fps}\\
   \midrule
SiamFC & 5.05 & 3.1 & 100\\
SiamRPN & 9.23 & 22.63 & 160\\
SiamRPN++ & 59.56 & 53.95 & 35\\
SiamBAN & 59.59 & 53.9 & 40\\
SiamTHN & 59.6 & 54.74 & 38\\
\bottomrule
\end{tabular}	
\end{table}

\textbf{Precision Plot}. The average euclidean distance between the center point predicted by the tracker and the ground truth center point is used to establish the central location error for each frame of the video. If this distance is below the specified threshold, the target is successfully tracked and the percentage value can be calculated by counting how many such frames there are. Different percentage values can be acquired depending on the threshold value, and therefore the precision plot can be obtained.

\textbf{Success Plot}. In each frame, $R_b$ denotes the bounding box predicted by the tracker and $R_{gt}$ represents the ground truth. We can calculate the size of the overlapping area between them by the following formula:
\begin{equation}
\label{deqn_ex6}
 OS = \frac{\left|R_b \cap R_{gt}\right|}{\left|R_b \cup R_{gt}\right|}.
\end{equation}
The percentage of frames where $OS$ is below the overlap threshold is the success rate. By setting different overlap threshold from 0 to 1, we can get a success plot. In addition, AUC is the area under the curve in the success plot, which can be used as a tracking accuracy evaluation metric.

Following the VOT evaluation protocols, VOT2016\cite{kristan2016novel} uses three evaluation metrics: Accuracy (A), Robustness (R), and Expected Average Overlap (EAO). A represents the average overlap between ground truth and the bounding box predicted by tracker during its successful tracking. R is used to evaluate the number of times the tracker loses a target during tracking. It is worthy to note that whenever a tracker loses the target object during the assessment, it is reset. EAO uses the raw data from A and R to estimate the average overlap predicted by the tracker over a huge number of short-term sequences that share the same visual characteristics with the given dataset.

\subsection{Implementation details}
Our approach is implemented under PyTorch 1.8.0 framework on a Intel(R) Xeon(R) Silver 4210R CPU(2.40GHz) along with a Nvidia Geforce RTX 3090GPU. The backbone is modified ResNet-50\cite{he2016deep}. And it is initialized using weights which is trained on ImageNet\cite{russakovsky2015imagenet}, and the parameters of the first two layers are frozen throughout training. Because the shallow layers of the network are more generalized and high layers are more related to specific tasks. GOT-10k\cite{huang2019got}, COCO\cite{lin2014microsoft}, ImageNet VID\cite{russakovsky2015imagenet}, and ImageNet DET\cite{russakovsky2015imagenet} provided the data for the training set. All of the images are cropped and scaled to 127×127 and 511×511 according to the ground truth provided in the dataset. We utilize corrective loss to train SiamTHN. The whole network is trained with 20 epcohs using Stochastic Gradient Descent (SGD) with a momentum of 0.9. Batch size is 28. Learning rate for the first 5 warm-up epochs varies from 0.001 to 0.005; for the following 15 epochs, it ranges from 0.005 to 0.00005.

\begin{table}[t]
\caption{ABLATION STUDY OF EFFECTIVENESS OF THM\label{tab:table3}}
\centering
\setlength{\tabcolsep}{1mm}{
\begin{tabular}{ccccc}
\toprule
   \textbf{Tracker} & Success Rate & Precision Rate & Params(M) & fps\\
   \midrule
SiamBAN & 0.543 & 0.713 & 53.9 & 40\\
SiamBAN + SE & 0.556 & 0.740 & 54.3 & 38\\
\textbf{SiamBAN + THM} & \textbf{0.582} & \textbf{0.762} & 54.74 & 38\\
\hline
SiamRPN++ & 0.528 & 0.696 & 53.95 & 35\\
\textbf{SiamRPN++ + THM} & \textbf{0.546} & \textbf{0.714} & 54.75 & 30\\
\hline
SiamCAR & 0.536 & 0.732 & 51.38 & 42\\
\textbf{SiamCAR + THM} & \textbf{0.556} & \textbf{0.737} & 52.18 & 39\\
\bottomrule
\end{tabular}}	
\end{table}

\subsection{Comparison on Public Benchmarks}
This section presents the tracking reults from our method and other trackers for OTB-2015, VOT2016, DTB70, UAV123 and UAV20L datasets, respectively.

\textbf{OTB-2015}\cite{7001050}. We evaluate our tracker against 9 state-of-the-art methods including AiATrack\cite{gao2022aiatrack}, TCTrack++\cite{cao2022tctrack}, TransT\cite{chen2021transformer}, Stark\cite{yan2021learning}, KYS\cite{bhat2020know}, SiamBAN\cite{chen2020siamese},  PrDiMP\cite{danelljan2020probabilistic}, SiamRPN\cite{li2018high} and DeepSRDCF\cite{danelljan2015convolutional}. As shown in Fig \ref{Baojh4}, our tracker is well prepared to handle challenging factors such as \textit{Background Clutters} and \textit{Out-of-Plane Rotation}. When faced with the \textit{Background Clutters} challenge, our tracker achieves a result of 0.674/0.886 on success plot and precision plot, outperforming existing trackers, which is a good proof that THM can effectively help similarity response map to focus on target region. When faced with the challenges of \textit{Out-of-Plane Rotation}, our tracker achieves 0.695/0.921, which is similar to the latest tracker AiATrack. Compared to the baseline tracker, our tracker improves the success rate by 0.013 and the precision rate by 0.023. The results show that our SiamTHN can better handle the background occlusion problem and scale change problem of the target, which benefit from our proposed Target Highlight Module and corrective loss.

\textbf{VOT2016}\cite{kristan2016novel}. Nearly all of the top-performing trackers from the VOT2016 are compared in Table~\ref{tab:table1}. As shown in the Table~\ref{tab:table1}, our tracker achieves the best EAO (0.510) and Robustnes (0.126) on the VOT2016 dataset. Compared to the baseline tracker, our tracker has a reduction of 0.24 on Robustnes. This shows that our tracker is able to have better robustness compared to the previous tracker while maintaining good tracking accuracy.

\textbf{DTB70}\cite{li2017visual}. Our tracker is compared to the 9 best performing trackers on DTB70, including SiamBAN\cite{chen2020siamese}, CLNet\cite{dong2020clnet}, TCTrack++\cite{cao2022tctrack}, TCTrack\cite{cao2022tctrack},  HiFT\cite{cao2021hift}, SiamRPN\cite{li2018high}, SiamFC\cite{bertinetto2016fully}, ECO\cite{danelljan2017eco} and STRCF\cite{li2018learning}. As shown in Fig \ref{Baojh5}, our tracker achieves 0.658 on success plot, which is slightly higher than the latest tracker, CLNet. Besides, it improves 0.016 compared to the baseline tracker (SiamBAN). This fully illustrates how effective our proposed tracker is.

\textbf{UAV123}\cite{mueller2016benchmark}. Our tracker is compared to the 9 best performing trackers on UAV123, including SiamRPN++\cite{li2019siamrpn++}, SiamBAN\cite{chen2020siamese}, CLNet\cite{dong2020clnet}, HiFT\cite{cao2021hift}, TCTrack++\cite{cao2022tctrack}, TCTrack\cite{cao2022tctrack}, SiamRPN\cite{li2018high}, ECO\cite{danelljan2017eco} and SRDCF\cite{danelljan2015learning}. As shown in Fig \ref{Baojh6}, our tracker achieves state-of-the-art scores of 0.635 and 0.836 on success plot and precision plot, which is similar to the latest tracker, CLNet.

\begin{figure}[t]
\centering
\includegraphics[width=3.5in]{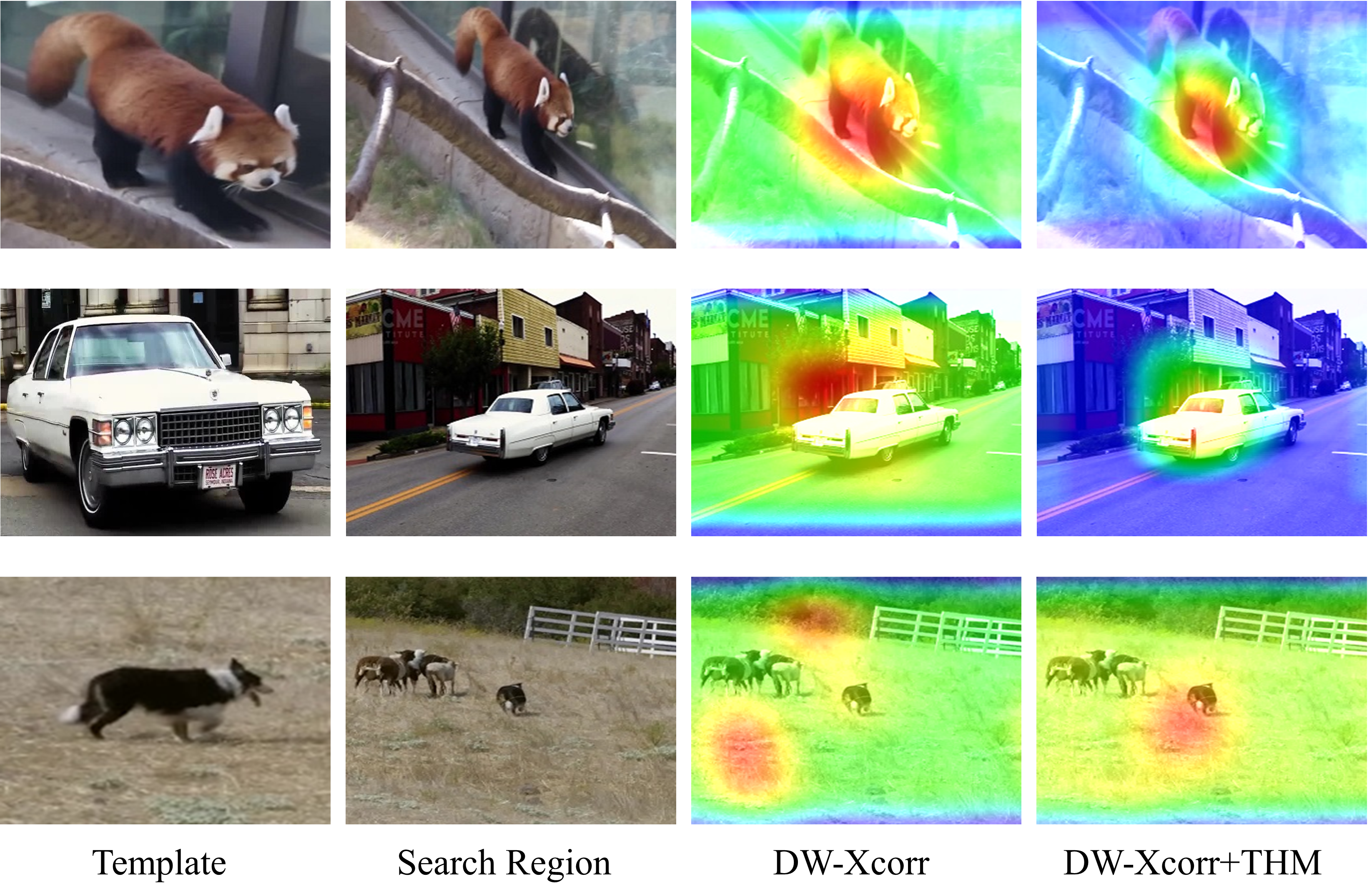}
\caption{Comparison between previous method (DW-Xcorr) and the our method (DW-Xcorr + THM) on example frames. In the previous method\cite{chen2020siamese}, the similarity response maps generated by DW-Xcorr do not accurately capture the target and are susceptible to the background. With the THM, similarity response map generated by our method is able to focus on regions belonging to the target.}
\label{Baojh8}
\end{figure}

\textbf{UAV20L}\cite{mueller2016benchmark}. Our tracker is compared to the 9 best performing trackers on UAV20L, including HiFT\cite{cao2021hift}, SiamAPN++\cite{cao2021siamapn++}, SiamAPN\cite{fu2021siamese}, SiamBAN\cite{chen2020siamese}, SiamRPN\cite{li2018high}, SiamFC\cite{bertinetto2016fully}, BACF\cite{kiani2017learning}, ECO\cite{danelljan2017eco} and STRCF\cite{li2018learning}. As shown in Fig \ref{Baojh7}, our tracker outperforms most other state-of-the-art trackers with scores of 0.594 and 0.776 on success plot and precision plot.

Overall, our tracker outperforms the competition on several tracking benchmark datasets, including the traditional target tracking datasets OTB-2015 and VOT2016, as well as the UAV aerial photography datasets DTB70, UAV123, and UAV20L. The above experimental findings clearly confirm the efficacy and generalizability of our model. At the same time, we have conducted an analysis of computational complexity, which can be seen in Table~\ref{tab:table2}. Compared with the baseline SiamBAN our method only increases 0.31 GFlops and 0.84 M in terms of computation and parameters. Our method also has an advantage in computational complexity compared to some other advanced trackers.

\begin{table}[t]
\caption{ABLATION STUDY OF EFFECTIVENESS OF CORRECTIVE LOSS\label{tab:table4}}
\centering
\setlength{\tabcolsep}{1mm}{
\begin{tabular}{ccc}
\toprule
   \textbf{Tracker} & Success Rate & Precision Rate\\
   \midrule
SiamBAN & 0.543 & 0.713\\
SiamBAN + iou-aware loss & 0.540 & 0.716\\
\textbf{SiamBAN + corrective loss} & \textbf{0.568} & \textbf{0.746}\\
\hline
SiamRPN++ & 0.528 & 0.696\\
\textbf{SiamRPN++ + corrective loss} & \textbf{0.539} & \textbf{0.726}\\
\hline
SiamTHN & 0.582 & 0.762\\
\textbf{SiamTHN + corrective loss} & \textbf{0.594} & \textbf{0.776}\\
\bottomrule
\end{tabular}}	
\end{table}

\begin{table}[t]
\caption{ABLATION STUDY OF EFFECTIVENESS OF OUR WORK\label{tab:table5}}
\centering
\setlength{\tabcolsep}{1mm}{
\begin{tabular}{ccccc}
\toprule
   \textbf{Tracker} & Success Rate & Precision Rate & Params(M) & fps\\
   \midrule
SiamBAN & 0.543 & 0.713 & 53.9 & 40\\
SiamBAN + THM & 0.582 & 0.762 & 54.74 & 38\\
SiamBAN + CL & 0.568 & 0.746 & 53.9 & 41\\
\textbf{SiamBAN + THM + CL} & \textbf{0.594} & \textbf{0.776} & 54.74 & 37\\
\bottomrule
\end{tabular}}
\end{table}

\begin{figure}[t]
\centering
\includegraphics[width=3.5in]{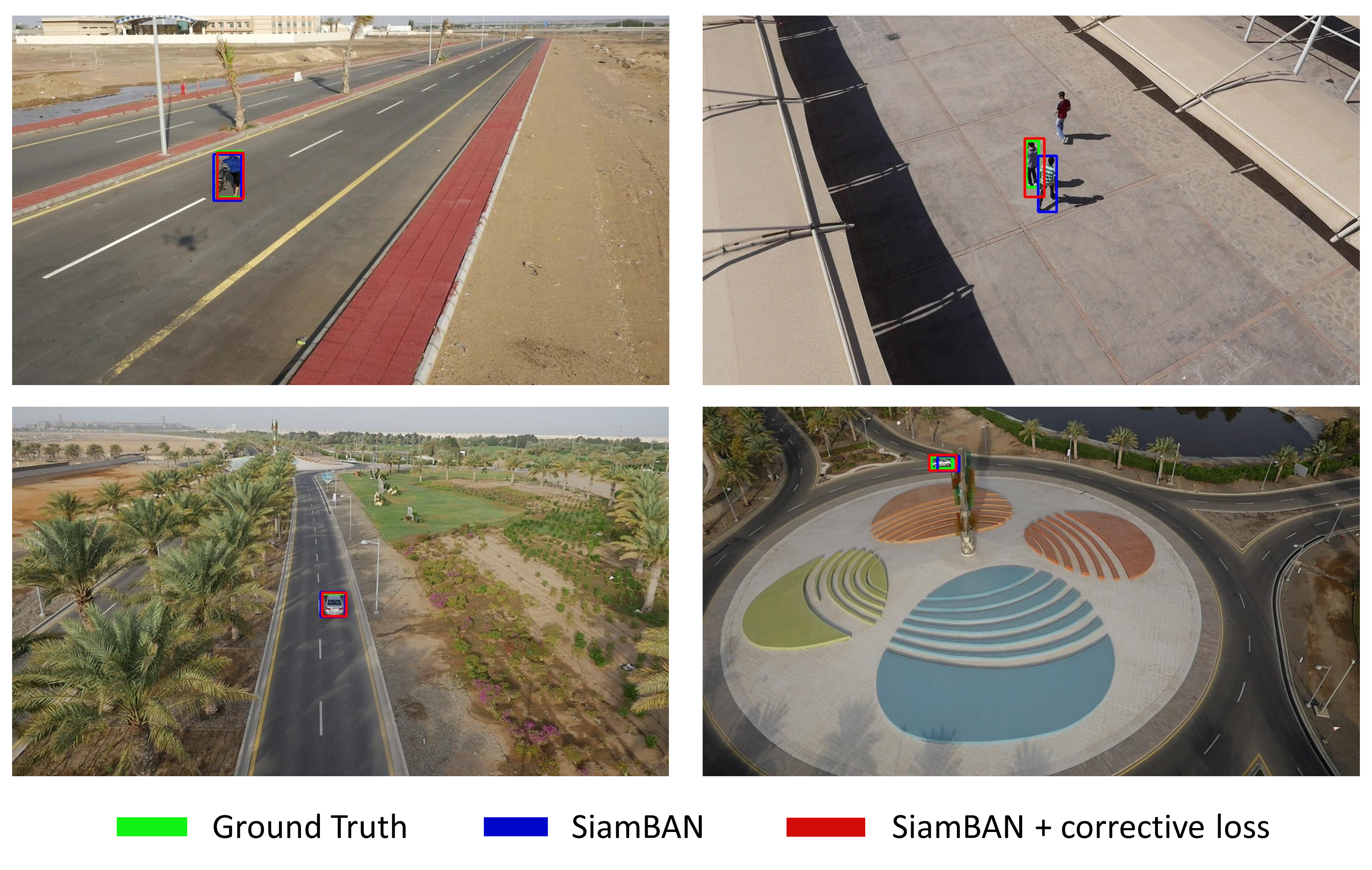}
\caption{Tracking comparison between SiamBAN and SiamBAN + corrective loss on example frames. The green bounding boxes denote the ground truth, while the tracking results produced by SiamBAN and SiamBAN + corrective loss are shown by the blue and red bounding boxes. The final generated tracking results are more accurate by using corrective loss to train SiamBAN.}
\label{Baojh9}
\end{figure}

\subsection{Ablation study}
In this section, we examine the impact of our Target Highlight Module and corrective loss, respectively, in order to illustrate the effectiveness of our SiamTHN. Furthermore, in order to demonstrate the superiority of our SiamTHN, it is compared with other state-of-the-art trackers in terms of each attribute of the UAV20L dataset. 

\begin{figure*}
\centering
\includegraphics[width=7in, height=9.5in]{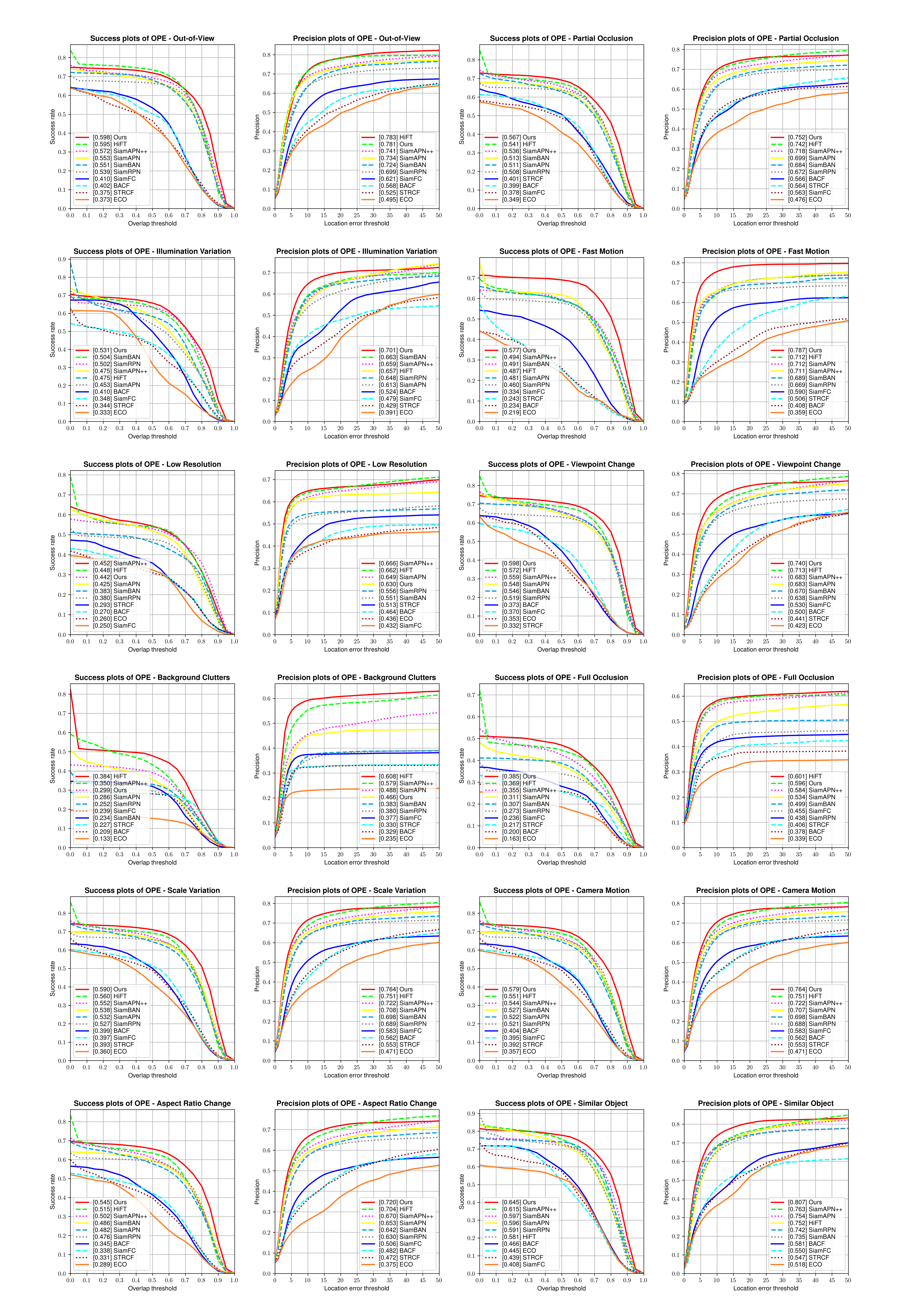}
\caption{Tracking output of attribute analysis on UAV20L benchmarks. UAV20L contains a total of 12 challenges, and we achieve state-of-the-art performance on Out-of-View, Partial Occlusion, IIIumination Variation, Fast Motion, Viewpoint Change, Full Occlusion, Scale Variation, Camera Motion, Aspect Ratio Change and Similar object.}
\label{Baojh10}
\end{figure*}

\begin{figure*}
\centering
\includegraphics[width=7in, height=4in]{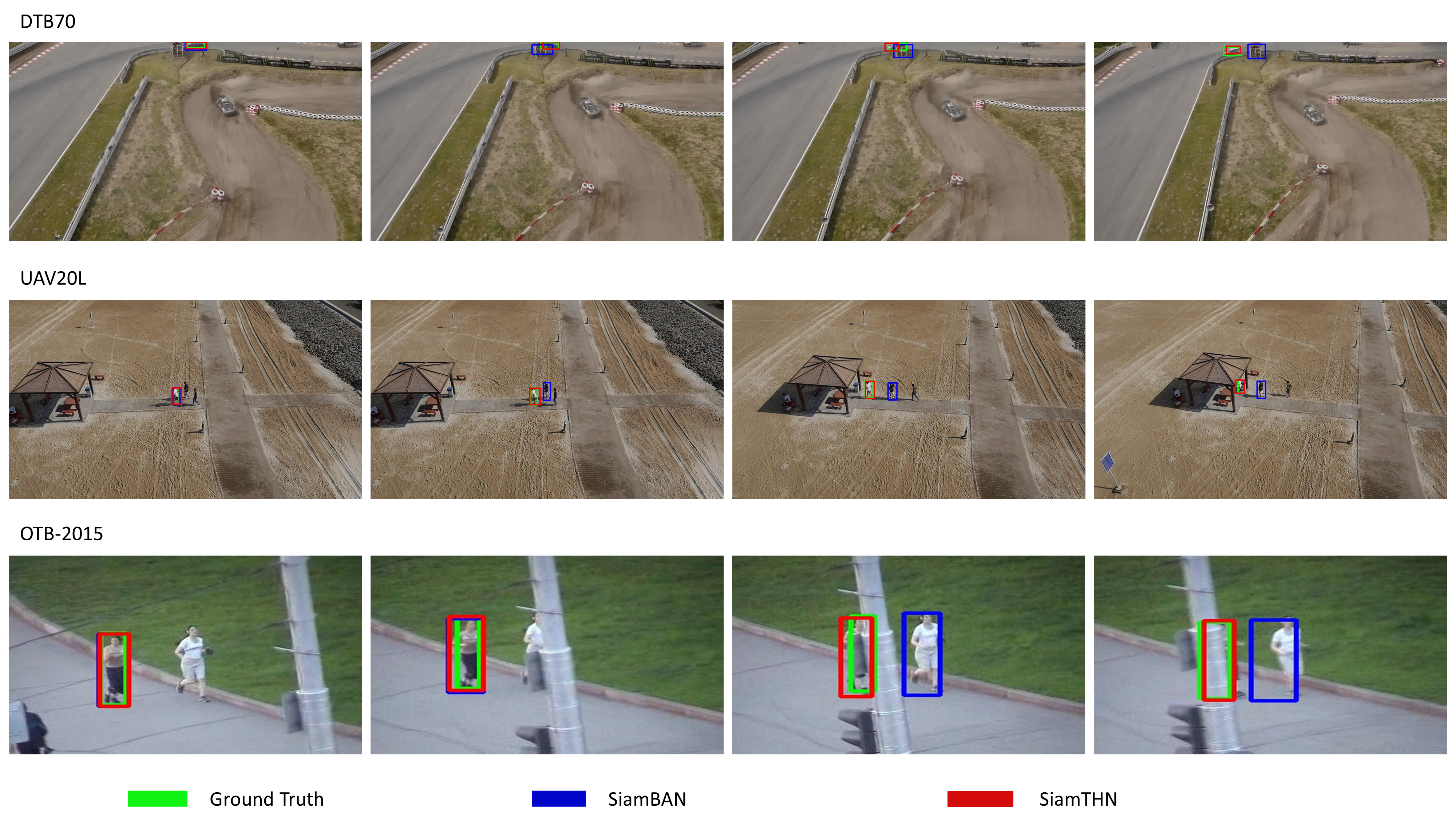}
\caption{Visualization of tracking results on videos from different datasets. The first row is DTB70 dataset, where the main challenge is small object. The second row is UAV20L dataset, where the main challenge is similar object. The third row is OTB-2015 dataset, where the main challenge is occlusion.}
\label{Baojh11}
\end{figure*}

\textbf{Analysis on SiamTHN.} To prove the effect of our SiamTHN, we carry four experiments using different models based on the same dataset and same hyperparameters configurations, which are as follows.

1) Baseline: To better perform the ablation experiments, we do not use the model provided by the authors of research, but reproduce SiamBAN using the four training datasets we utilized. The performance of the SiamBAN model reproduced is then evaluated using the UAV20L dataset. As shown in Table~\ref{tab:table5}, SiamBAN achieves 0.543/0.713 on success plot and precision plot.

2) Improvement of THM: We use the Target Highlight Module(THM) to enhance the feature map's presentation of the target and further enabling the similarity response map to better focus on the target region. As shown in Table~\ref{tab:table3}, the tracking performance increases to 0.582/0.762 on success plot and precision plot once our Target Highlight Module is integrated into SiamBAN. THM helps to improve the success rate of 0.039 and the accuracy of 0.049 without adding too many extra parameters and computations.When we add the traditional attention mechanism to the similarity matching module, it achieves 0.556/0.740 on success plot and precision plot. It helps to improve the success rate of 0.013 and the accuracy of 0.027, whcih is much lower results than our THM. Because the traditional attention mechanism does not focus on our concerns. Finally, we add THM to the other two methods and compare them with the original methods comprehensively to demonstrate the effectiveness of THM. As shown in Table~\ref{tab:table3}, with the help of THM, the performance of both SiamRPN++\cite{li2019siamrpn++} and SiamCAR\cite{guo2020siamcar} is all improved. The success rate of SiamRPN++ is increased by 0.018, and the precision rate is increased by 0.018. The success rate of SiamCAR is increased by 0.02, and the precision rate is increased by 0.005. And the addition of THM only increases the number of parameters by a small amount and their tracking speed hardly decreases.

To confirm the effectiveness of the Target Highlight Module(THM), we compare the similarity response maps produced by DW-Xcorr and THM + DW-Xcorr, respectively. As shown in Fig \ref{Baojh8}, we show three different scenarios. In the first line, THM can help similarity response map has high responsiveness to the target region, and low responsiveness at the rest of the locations. In the second line, THM makes the point with the highest response value in the similarity response map more concentrated in the target region. In the third line, otherwise misdirected response points can be refocused on the target region with the help of THM. This comparison figure fully demonstrates that THM can make the similarity response map generated by DW-Xcorr more focused on the target region and thus produce more accurate prediction results.

3) Improvement of corrective loss: Corrective loss optimizes both the classification and regression branches together, resulting in more consistent classification scores and regression bounding boxes.  We use corrective loss to train SiamBAN, and the tracking results are improved, reaching 0.568/0.746 on success plot and precision plot. As shown in Table~\ref{tab:table4}, corrective loss helps to improve the success rate of 0.025 and the precision rate of 0.033. In addition, we set up a set of experiments to illustrate the superiority of our corrective loss compared to the previous loss function. In order to fairly compare the effectiveness of the loss functions, we do not introduce additionaul branches. As shown in Table~\ref{tab:table4}, using iou-aware loss alone does not improve the training effect of the model. It requires additional localization branches to be effective. To better demonstrate the expansibility of corrective loss, we also apply it to the training process of multiple trackers. As shown in Table~\ref{tab:table4}, training SiamRPN++ and SiamTHN with corrective loss can further improve their performance. The success rate of SiamRPN++ is increased by 0.011, and the precision rate is increased by 0.03. The success rate of SiamTHN is increased by 0.012, and the precision rate is increased by 0.014.

We compare the tracking results produced by SiamBAN and SiamBAN + corrective loss, respectively, to confirm the effectiveness of the corrective loss. To show the effect of corrective loss more visually, we also output the tracking results of SiamBAN and SiamBAN + corrective loss. As shown in Fig \ref{Baojh9}, using corrective loss to train SiamBAN, the final model obtained generates a more accurate tracking frame. This is strong evidence for the effectiveness of corrective loss.

4) SiamBAN + Target Highlight Module + corrective loss: As shown in Table~\ref{tab:table5}, we use SiamBAN as the baseline to set up ablation experiments about Target Highlight Module(THM) and corrective loss. With the help of both THM and corrective loss, the success rate and precision rate finally are increased to 0.594/0.766 on success rate and precision rate, which achieve state-of-the-art score. In addition, we compare our tracker with other state-of-the-art trackers in terms of each attribute of the UAV20L dataset. As shown in Fig \ref{Baojh10}, our tracker performs best on most attributes, including \textit{out-of-view}, \textit{scale rariation}, \textit{spect ratio change} and \textit{similar object}. This indicates the strong robustness of our model in the face of multiple challenges.

\subsection{Visualization.}
To show our tracker's superiority, we compare its performance on several datasets with that of the baseline tracker. As shown in Fig \ref{Baojh11}, we perform validation on three datasets, DTB70, UAV20L, and OTB-2015, and select some challenging scenarios for visualization. Our tracker is more robust to challenging factors such as small object, similar object and occlusion. For example, the challenges faced in the first row are small object, the second row is a similar object, and the third row is occlusion. Our tracker is able to accurately locate the target due to the utilize of THM and corrective loss.

\section{Conclusion}
In this paper, we propose a siamese network framework for efficient target tracking. Specifically, we propose a Target Highlight Module to adaptively balance the weights among different channels to obtain more representative output features, which makes the similarity response maps generated by DW-Xcorr more focused on the target region. Furthermore, we propose to train the model using corrective loss to optimize both classification and regression branches, eliminating the misalignment between classification branch and regression branch. Experimental results on five tracking benchmarks shows that our proposed Siamese Target Highlight Network (SiamTHN) achieves state-of-the-art performance, running at 38 frames per second on a Nvidia RTX 3090, confirming its effectiveness and efficiency. In particular, our tracker performs better than the existing tracker when faced with small object, similar object and occlusion.

\bibliographystyle{IEEEtran}
\bibliography{IEEEfull}

% Generated by IEEEtran.bst, version: 1.12 (2007/01/11)
\begin{thebibliography}{10}
\providecommand{\url}[1]{#1}
\csname url@samestyle\endcsname
\providecommand{\newblock}{\relax}
\providecommand{\bibinfo}[2]{#2}
\providecommand{\BIBentrySTDinterwordspacing}{\spaceskip=0pt\relax}
\providecommand{\BIBentryALTinterwordstretchfactor}{4}
\providecommand{\BIBentryALTinterwordspacing}{\spaceskip=\fontdimen2\font plus
\BIBentryALTinterwordstretchfactor\fontdimen3\font minus
  \fontdimen4\font\relax}
\providecommand{\BIBforeignlanguage}[2]{{%
\expandafter\ifx\csname l@#1\endcsname\relax
\typeout{** WARNING: IEEEtran.bst: No hyphenation pattern has been}%
\typeout{** loaded for the language `#1'. Using the pattern for}%
\typeout{** the default language instead.}%
\else
\language=\csname l@#1\endcsname
\fi
#2}}
\providecommand{\BIBdecl}{\relax}
\BIBdecl

\bibitem{kart2019object}
U.~Kart, A.~Lukezic, M.~Kristan, J.-K. Kamarainen, and J.~Matas, ``Object
  tracking by reconstruction with view-specific discriminative correlation
  filters,'' in \emph{Proceedings of the IEEE/CVF Conference on Computer Vision
  and Pattern Recognition}, 2019, pp. 1339--1348.

\bibitem{shi2020adaptive}
Y.~Shi, Z.~Wei, H.~Ling, Z.~Wang, P.~Zhu, J.~Shen, and P.~Li, ``Adaptive and
  robust partition learning for person retrieval with policy gradient,''
  \emph{IEEE Transactions on Multimedia}, vol.~23, pp. 3264--3277, 2020.

\bibitem{wu2022pseudo}
L.~Wu, D.~Liu, W.~Zhang, D.~Chen, Z.~Ge, F.~Boussaid, M.~Bennamoun, and
  J.~Shen, ``Pseudo-pair based self-similarity learning for unsupervised person
  re-identification,'' \emph{IEEE Transactions on Image Processing}, vol.~31,
  pp. 4803--4816, 2022.

\bibitem{gao2019manifold}
M.~Gao, L.~Jin, Y.~Jiang, and B.~Guo, ``Manifold siamese network: A novel
  visual tracking convnet for autonomous vehicles,'' \emph{IEEE Transactions on
  Intelligent Transportation Systems}, vol.~21, no.~4, pp. 1612--1623, 2019.

\bibitem{henriques2014high}
J.~F. Henriques, R.~Caseiro, P.~Martins, and J.~Batista, ``High-speed tracking
  with kernelized correlation filters,'' \emph{IEEE transactions on pattern
  analysis and machine intelligence}, vol.~37, no.~3, pp. 583--596, 2014.

\bibitem{li2014scale}
Y.~Li and J.~Zhu, ``A scale adaptive kernel correlation filter tracker with
  feature integration,'' in \emph{European conference on computer
  vision}.\hskip 1em plus 0.5em minus 0.4em\relax Springer, 2014, pp. 254--265.

\bibitem{kiani2017learning}
H.~Kiani~Galoogahi, A.~Fagg, and S.~Lucey, ``Learning background-aware
  correlation filters for visual tracking,'' in \emph{Proceedings of the IEEE
  international conference on computer vision}, 2017, pp. 1135--1143.

\bibitem{han2018adaptive}
Z.~Han, P.~Wang, and Q.~Ye, ``Adaptive discriminative deep correlation filter
  for visual object tracking,'' \emph{IEEE Transactions on Circuits and Systems
  for Video Technology}, vol.~30, no.~1, pp. 155--166, 2018.

\bibitem{zhu2020complementary}
X.-F. Zhu, X.-J. Wu, T.~Xu, Z.-H. Feng, and J.~Kittler, ``Complementary
  discriminative correlation filters based on collaborative representation for
  visual object tracking,'' \emph{IEEE Transactions on Circuits and Systems for
  Video Technology}, vol.~31, no.~2, pp. 557--568, 2020.

\bibitem{jain2021channel}
M.~Jain, A.~Tyagi, A.~V. Subramanyam, S.~Denman, S.~Sridharan, and C.~Fookes,
  ``Channel graph regularized correlation filters for visual object tracking,''
  \emph{IEEE Transactions on Circuits and Systems for Video Technology},
  vol.~32, no.~2, pp. 715--729, 2021.

\bibitem{bertinetto2016fully}
L.~Bertinetto, J.~Valmadre, J.~F. Henriques, A.~Vedaldi, and P.~H. Torr,
  ``Fully-convolutional siamese networks for object tracking,'' in
  \emph{European conference on computer vision}.\hskip 1em plus 0.5em minus
  0.4em\relax Springer, 2016, pp. 850--865.

\bibitem{li2019siamrpn++}
B.~Li, W.~Wu, Q.~Wang, F.~Zhang, J.~Xing, and J.~Yan, ``Siamrpn++: Evolution of
  siamese visual tracking with very deep networks,'' in \emph{Proceedings of
  the IEEE/CVF Conference on Computer Vision and Pattern Recognition}, 2019,
  pp. 4282--4291.

\bibitem{chen2020siamese}
Z.~Chen, B.~Zhong, G.~Li, S.~Zhang, and R.~Ji, ``Siamese box adaptive network
  for visual tracking,'' in \emph{Proceedings of the IEEE/CVF conference on
  computer vision and pattern recognition}, 2020, pp. 6668--6677.

\bibitem{fan2020feature}
J.~Fan, H.~Song, K.~Zhang, K.~Yang, and Q.~Liu, ``Feature alignment and
  aggregation siamese networks for fast visual tracking,'' \emph{IEEE
  Transactions on Circuits and Systems for Video Technology}, vol.~31, no.~4,
  pp. 1296--1307, 2020.

\bibitem{jiang2020mutual}
M.~Jiang, Y.~Zhao, and J.~Kong, ``Mutual learning and feature fusion siamese
  networks for visual object tracking,'' \emph{IEEE Transactions on Circuits
  and Systems for Video Technology}, vol.~31, no.~8, pp. 3154--3167, 2020.

\bibitem{fan2021siamon}
C.~Fan, H.~Yu, Y.~Huang, C.~Shan, L.~Wang, and C.~Li, ``Siamon: Siamese
  occlusion-aware network for visual tracking,'' \emph{IEEE Transactions on
  Circuits and Systems for Video Technology}, 2021.

\bibitem{wang2021dynamic}
X.~Wang, Z.~Chen, J.~Tang, B.~Luo, Y.~Wang, Y.~Tian, and F.~Wu, ``Dynamic
  attention guided multi-trajectory analysis for single object tracking,''
  \emph{IEEE Transactions on Circuits and Systems for Video Technology},
  vol.~31, no.~12, pp. 4895--4908, 2021.

\bibitem{bromley1993signature}
J.~Bromley, I.~Guyon, Y.~LeCun, E.~S{\"a}ckinger, and R.~Shah, ``Signature
  verification using a" siamese" time delay neural network,'' \emph{Advances in
  neural information processing systems}, vol.~6, 1993.

\bibitem{guo2017learning}
Q.~Guo, W.~Feng, C.~Zhou, R.~Huang, L.~Wan, and S.~Wang, ``Learning dynamic
  siamese network for visual object tracking,'' in \emph{Proceedings of the
  IEEE international conference on computer vision}, 2017, pp. 1763--1771.

\bibitem{wang2018not}
Q.~Wang, M.~Zhang, J.~Xing, J.~Gao, W.~Hu, and S.~J. Maybank, ``Do not lose the
  details: reinforced representation learning for high performance visual
  tracking,'' in \emph{27th International Joint Conference on Artificial
  Intelligence}, 2018.

\bibitem{he2018twofold}
A.~He, C.~Luo, X.~Tian, and W.~Zeng, ``A twofold siamese network for real-time
  object tracking,'' in \emph{Proceedings of the IEEE conference on computer
  vision and pattern recognition}, 2018, pp. 4834--4843.

\bibitem{li2018high}
B.~Li, J.~Yan, W.~Wu, Z.~Zhu, and X.~Hu, ``High performance visual tracking
  with siamese region proposal network,'' in \emph{Proceedings of the IEEE
  conference on computer vision and pattern recognition}, 2018, pp. 8971--8980.

\bibitem{ren2015faster}
S.~Ren, K.~He, R.~Girshick, and J.~Sun, ``Faster r-cnn: Towards real-time
  object detection with region proposal networks,'' \emph{Advances in neural
  information processing systems}, vol.~28, 2015.

\bibitem{guo2020siamcar}
D.~Guo, J.~Wang, Y.~Cui, Z.~Wang, and S.~Chen, ``Siamcar: Siamese fully
  convolutional classification and regression for visual tracking,'' in
  \emph{Proceedings of the IEEE/CVF conference on computer vision and pattern
  recognition}, 2020, pp. 6269--6277.

\bibitem{vaswani2017attention}
A.~Vaswani, N.~Shazeer, N.~Parmar, J.~Uszkoreit, L.~Jones, A.~N. Gomez,
  {\L}.~Kaiser, and I.~Polosukhin, ``Attention is all you need,''
  \emph{Advances in neural information processing systems}, vol.~30, 2017.

\bibitem{chen2021transformer}
X.~Chen, B.~Yan, J.~Zhu, D.~Wang, X.~Yang, and H.~Lu, ``Transformer tracking,''
  in \emph{Proceedings of the IEEE/CVF Conference on Computer Vision and
  Pattern Recognition}, 2021, pp. 8126--8135.

\bibitem{yan2021learning}
B.~Yan, H.~Peng, J.~Fu, D.~Wang, and H.~Lu, ``Learning spatio-temporal
  transformer for visual tracking,'' in \emph{Proceedings of the IEEE/CVF
  International Conference on Computer Vision}, 2021, pp. 10\,448--10\,457.

\bibitem{gao2022aiatrack}
S.~Gao, C.~Zhou, C.~Ma, X.~Wang, and J.~Yuan, ``Aiatrack: Attention in
  attention for transformer visual tracking,'' \emph{arXiv preprint
  arXiv:2207.09603}, 2022.

\bibitem{li2018deep}
P.~Li, D.~Wang, L.~Wang, and H.~Lu, ``Deep visual tracking: Review and
  experimental comparison,'' \emph{Pattern Recognition}, vol.~76, pp. 323--338,
  2018.

\bibitem{krizhevsky2012imagenet}
A.~Krizhevsky, I.~Sutskever, and G.~E. Hinton, ``Imagenet classification with
  deep convolutional neural networks,'' \emph{Advances in neural information
  processing systems}, vol.~25, 2012.

\bibitem{he2016deep}
K.~He, X.~Zhang, S.~Ren, and J.~Sun, ``Deep residual learning for image
  recognition,'' in \emph{Proceedings of the IEEE conference on computer vision
  and pattern recognition}, 2016, pp. 770--778.

\bibitem{long2015fully}
J.~Long, E.~Shelhamer, and T.~Darrell, ``Fully convolutional networks for
  semantic segmentation,'' in \emph{Proceedings of the IEEE conference on
  computer vision and pattern recognition}, 2015, pp. 3431--3440.

\bibitem{li2019gradnet}
P.~Li, B.~Chen, W.~Ouyang, D.~Wang, X.~Yang, and H.~Lu, ``Gradnet:
  Gradient-guided network for visual object tracking,'' in \emph{Proceedings of
  the IEEE/CVF International conference on computer vision}, 2019, pp.
  6162--6171.

\bibitem{fan2019siamese}
H.~Fan and H.~Ling, ``Siamese cascaded region proposal networks for real-time
  visual tracking,'' in \emph{Proceedings of the IEEE/CVF conference on
  computer vision and pattern recognition}, 2019, pp. 7952--7961.

\bibitem{xu2020siamfc++}
Y.~Xu, Z.~Wang, Z.~Li, Y.~Yuan, and G.~Yu, ``Siamfc++: Towards robust and
  accurate visual tracking with target estimation guidelines,'' in
  \emph{Proceedings of the AAAI Conference on Artificial Intelligence},
  vol.~34, no.~07, 2020, pp. 12\,549--12\,556.

\bibitem{zhang2020ocean}
Z.~Zhang, H.~Peng, J.~Fu, B.~Li, and W.~Hu, ``Ocean: Object-aware anchor-free
  tracking,'' in \emph{European Conference on Computer Vision}.\hskip 1em plus
  0.5em minus 0.4em\relax Springer, 2020, pp. 771--787.

\bibitem{ramesh2020tld}
B.~Ramesh, S.~Zhang, H.~Yang, A.~Ussa, M.~Ong, G.~Orchard, and C.~Xiang,
  ``e-tld: Event-based framework for dynamic object tracking,'' \emph{IEEE
  Transactions on Circuits and Systems for Video Technology}, vol.~31, no.~10,
  pp. 3996--4006, 2020.

\bibitem{chen2017sca}
L.~Chen, H.~Zhang, J.~Xiao, L.~Nie, J.~Shao, W.~Liu, and T.-S. Chua, ``Sca-cnn:
  Spatial and channel-wise attention in convolutional networks for image
  captioning,'' in \emph{Proceedings of the IEEE conference on computer vision
  and pattern recognition}, 2017, pp. 5659--5667.

\bibitem{hu2018squeeze}
J.~Hu, L.~Shen, and G.~Sun, ``Squeeze-and-excitation networks,'' in
  \emph{Proceedings of the IEEE conference on computer vision and pattern
  recognition}, 2018, pp. 7132--7141.

\bibitem{gao2019global}
Z.~Gao, J.~Xie, Q.~Wang, and P.~Li, ``Global second-order pooling convolutional
  networks,'' in \emph{Proceedings of the IEEE/CVF Conference on Computer
  Vision and Pattern Recognition}, 2019, pp. 3024--3033.

\bibitem{2020ECA}
Q.~Wang, B.~Wu, P.~Zhu, P.~Li, and Q.~Hu, ``Eca-net: Efficient channel
  attention for deep convolutional neural networks,'' in \emph{2020 IEEE/CVF
  Conference on Computer Vision and Pattern Recognition (CVPR)}, 2020.

\bibitem{lee2019srm}
H.~Lee, H.-E. Kim, and H.~Nam, ``Srm: A style-based recalibration module for
  convolutional neural networks,'' in \emph{Proceedings of the IEEE/CVF
  International Conference on Computer Vision}, 2019, pp. 1854--1862.

\bibitem{liu2016ssd}
W.~Liu, D.~Anguelov, D.~Erhan, C.~Szegedy, S.~Reed, C.-Y. Fu, and A.~C. Berg,
  ``Ssd: Single shot multibox detector,'' in \emph{European conference on
  computer vision}.\hskip 1em plus 0.5em minus 0.4em\relax Springer, 2016, pp.
  21--37.

\bibitem{girshick2015fast}
R.~Girshick, ``Fast r-cnn,'' in \emph{Proceedings of the IEEE international
  conference on computer vision}, 2015, pp. 1440--1448.

\bibitem{jiang2018acquisition}
B.~Jiang, R.~Luo, J.~Mao, T.~Xiao, and Y.~Jiang, ``Acquisition of localization
  confidence for accurate object detection,'' in \emph{Proceedings of the
  European conference on computer vision (ECCV)}, 2018, pp. 784--799.

\bibitem{cao2020prime}
Y.~Cao, K.~Chen, C.~C. Loy, and D.~Lin, ``Prime sample attention in object
  detection,'' in \emph{Proceedings of the IEEE/CVF Conference on Computer
  Vision and Pattern Recognition}, 2020, pp. 11\,583--11\,591.

\bibitem{wang2021reconcile}
K.~Wang and L.~Zhang, ``Reconcile prediction consistency for balanced object
  detection,'' in \emph{Proceedings of the IEEE/CVF International Conference on
  Computer Vision}, 2021, pp. 3631--3640.

\bibitem{tian2019fcos}
Z.~Tian, C.~Shen, H.~Chen, and T.~He, ``Fcos: Fully convolutional one-stage
  object detection,'' in \emph{Proceedings of the IEEE/CVF international
  conference on computer vision}, 2019, pp. 9627--9636.

\bibitem{peng2021siamrcr}
J.~Peng, Z.~Jiang, Y.~Gu, Y.~Wu, Y.~Wang, Y.~Tai, C.~Wang, and W.~Lin,
  ``Siamrcr: Reciprocal classification and regression for visual object
  tracking,'' \emph{arXiv preprint arXiv:2105.11237}, 2021.

\bibitem{yu2016unitbox}
J.~Yu, Y.~Jiang, Z.~Wang, Z.~Cao, and T.~Huang, ``Unitbox: An advanced object
  detection network,'' in \emph{Proceedings of the 24th ACM international
  conference on Multimedia}, 2016, pp. 516--520.

\bibitem{7001050}
Y.~Wu, J.~Lim, and M.-H. Yang, ``Object tracking benchmark,'' \emph{IEEE
  Transactions on Pattern Analysis and Machine Intelligence}, vol.~37, no.~9,
  pp. 1834--1848, 2015.

\bibitem{kristan2016novel}
M.~Kristan, J.~Matas, A.~Leonardis, T.~Voj{\'\i}{\v{r}}, R.~Pflugfelder,
  G.~Fernandez, G.~Nebehay, F.~Porikli, and L.~{\v{C}}ehovin, ``A novel
  performance evaluation methodology for single-target trackers,'' \emph{IEEE
  transactions on pattern analysis and machine intelligence}, vol.~38, no.~11,
  pp. 2137--2155, 2016.

\bibitem{li2017visual}
S.~Li and D.-Y. Yeung, ``Visual object tracking for unmanned aerial vehicles: A
  benchmark and new motion models,'' in \emph{Thirty-first AAAI conference on
  artificial intelligence}, 2017.

\bibitem{mueller2016benchmark}
M.~Mueller, N.~Smith, and B.~Ghanem, ``A benchmark and simulator for uav
  tracking,'' in \emph{European conference on computer vision}.\hskip 1em plus
  0.5em minus 0.4em\relax Springer, 2016, pp. 445--461.

\bibitem{huang2019got}
L.~Huang, X.~Zhao, and K.~Huang, ``Got-10k: A large high-diversity benchmark
  for generic object tracking in the wild,'' \emph{IEEE Transactions on Pattern
  Analysis and Machine Intelligence}, vol.~43, no.~5, pp. 1562--1577, 2019.

\bibitem{lin2014microsoft}
T.-Y. Lin, M.~Maire, S.~Belongie, J.~Hays, P.~Perona, D.~Ramanan,
  P.~Doll{\'a}r, and C.~L. Zitnick, ``Microsoft coco: Common objects in
  context,'' in \emph{European conference on computer vision}.\hskip 1em plus
  0.5em minus 0.4em\relax Springer, 2014, pp. 740--755.

\bibitem{russakovsky2015imagenet}
O.~Russakovsky, J.~Deng, H.~Su, J.~Krause, S.~Satheesh, S.~Ma, Z.~Huang,
  A.~Karpathy, A.~Khosla, M.~Bernstein \emph{et~al.}, ``Imagenet large scale
  visual recognition challenge,'' \emph{International journal of computer
  vision}, vol. 115, no.~3, pp. 211--252, 2015.

\bibitem{miller1995wordnet}
G.~A. Miller, ``Wordnet: a lexical database for english,'' \emph{Communications
  of the ACM}, vol.~38, no.~11, pp. 39--41, 1995.

\bibitem{danelljan2017eco}
M.~Danelljan, G.~Bhat, F.~Shahbaz~Khan, and M.~Felsberg, ``Eco: Efficient
  convolution operators for tracking,'' in \emph{Proceedings of the IEEE
  conference on computer vision and pattern recognition}, 2017, pp. 6638--6646.

\bibitem{danelljan2019atom}
M.~Danelljan, G.~Bhat, F.~S. Khan, and M.~Felsberg, ``Atom: Accurate tracking
  by overlap maximization,'' in \emph{Proceedings of the IEEE/CVF Conference on
  Computer Vision and Pattern Recognition}, 2019, pp. 4660--4669.

\bibitem{voigtlaender2020siam}
P.~Voigtlaender, J.~Luiten, P.~H. Torr, and B.~Leibe, ``Siam r-cnn: Visual
  tracking by re-detection,'' in \emph{Proceedings of the IEEE/CVF Conference
  on Computer Vision and Pattern Recognition}, 2020, pp. 6578--6588.

\bibitem{danelljan2020probabilistic}
M.~Danelljan, L.~V. Gool, and R.~Timofte, ``Probabilistic regression for visual
  tracking,'' in \emph{Proceedings of the IEEE/CVF conference on computer
  vision and pattern recognition}, 2020, pp. 7183--7192.

\bibitem{cao2022tctrack}
Z.~Cao, Z.~Huang, L.~Pan, S.~Zhang, Z.~Liu, and C.~Fu, ``Tctrack: Temporal
  contexts for aerial tracking,'' in \emph{Proceedings of the IEEE/CVF
  Conference on Computer Vision and Pattern Recognition}, 2022, pp.
  14\,798--14\,808.

\bibitem{bhat2020know}
G.~Bhat, M.~Danelljan, L.~V. Gool, and R.~Timofte, ``Know your surroundings:
  Exploiting scene information for object tracking,'' in \emph{European
  Conference on Computer Vision}.\hskip 1em plus 0.5em minus 0.4em\relax
  Springer, 2020, pp. 205--221.

\bibitem{danelljan2015convolutional}
M.~Danelljan, G.~Hager, F.~Shahbaz~Khan, and M.~Felsberg, ``Convolutional
  features for correlation filter based visual tracking,'' in \emph{Proceedings
  of the IEEE international conference on computer vision workshops}, 2015, pp.
  58--66.

\bibitem{dong2020clnet}
X.~Dong, J.~Shen, L.~Shao, and F.~Porikli, ``Clnet: A compact latent network
  for fast adjusting siamese trackers,'' in \emph{European Conference on
  Computer Vision}.\hskip 1em plus 0.5em minus 0.4em\relax Springer, 2020, pp.
  378--395.

\bibitem{cao2021hift}
Z.~Cao, C.~Fu, J.~Ye, B.~Li, and Y.~Li, ``Hift: Hierarchical feature
  transformer for aerial tracking,'' in \emph{Proceedings of the IEEE/CVF
  International Conference on Computer Vision}, 2021, pp. 15\,457--15\,466.

\bibitem{li2018learning}
F.~Li, C.~Tian, W.~Zuo, L.~Zhang, and M.-H. Yang, ``Learning spatial-temporal
  regularized correlation filters for visual tracking,'' in \emph{Proceedings
  of the IEEE conference on computer vision and pattern recognition}, 2018, pp.
  4904--4913.

\bibitem{danelljan2015learning}
M.~Danelljan, G.~Hager, F.~Shahbaz~Khan, and M.~Felsberg, ``Learning spatially
  regularized correlation filters for visual tracking,'' in \emph{Proceedings
  of the IEEE international conference on computer vision}, 2015, pp.
  4310--4318.

\bibitem{cao2021siamapn++}
Z.~Cao, C.~Fu, J.~Ye, B.~Li, and Y.~Li, ``Siamapn++: Siamese attentional
  aggregation network for real-time uav tracking,'' in \emph{2021 IEEE/RSJ
  International Conference on Intelligent Robots and Systems (IROS)}.\hskip 1em
  plus 0.5em minus 0.4em\relax IEEE, 2021, pp. 3086--3092.

\bibitem{fu2021siamese}
C.~Fu, Z.~Cao, Y.~Li, J.~Ye, and C.~Feng, ``Siamese anchor proposal network for
  high-speed aerial tracking,'' in \emph{2021 IEEE International Conference on
  Robotics and Automation (ICRA)}.\hskip 1em plus 0.5em minus 0.4em\relax IEEE,
  2021, pp. 510--516.

\end{thebibliography}

\end{document}